\documentclass[iicol]{sn-jnl}

\usepackage{graphicx}%
\usepackage{multirow}%
\usepackage{amsmath,amssymb,amsfonts}%
\usepackage{amsthm}%
\usepackage{mathrsfs}%
\usepackage[title]{appendix}%
\usepackage{xcolor}%
\usepackage{textcomp}%
\usepackage{manyfoot}%
\usepackage{booktabs}%
\usepackage{algorithm}%
\usepackage{algorithmicx}%
\usepackage{algpseudocode}%
\usepackage{listings}%

\usepackage{natbib}
\usepackage{pifont}

\begin{document}

\title[Test-Time Computing for Referring Multimodal Large Language Models]{Test-Time Computing for Referring Multimodal Large Language Models}


\author[1,4]{\fnm{Mingrui} \sur{Wu}}
\equalcont{These authors contributed equally to this work.}

\author[1]{\fnm{Hao} \sur{Chen}}
\equalcont{These authors contributed equally to this work.}

\author[1]{\fnm{Jiayi} \sur{Ji}}
\equalcont{These authors contributed equally to this work.}

\author[1]{\fnm{Xiaoshuai} \sur{Sun}}

\author[3]{\fnm{Zhiyuan} \sur{Liu}}

\author[1]{\fnm{Liujuan} \sur{Cao}}

\author[2,4]{\fnm{Ming-Ming} \sur{Cheng}}

\author*[1]{\fnm{Rongrong} \sur{Ji}}\email{rrji@xmu.edu.cn}

\affil[1]{Key Laboratory of Multimedia Trusted Perception and Efficient Computing, Ministry of Education of China, Xiamen University, 361005, P.R. China}

\affil[2]{VCIP, CS, Nankai University}

\affil[3]{Tsinghua University}
\affil[4]{Zhongguancun Academy, Beijing, China.100094}


\abstract{We propose \textbf{ControlMLLM++}, a novel test-time adaptation framework that injects learnable visual prompts into frozen multimodal large language models (MLLMs) to enable fine-grained region-based visual reasoning without any model retraining or fine-tuning. Leveraging the insight that cross-modal attention maps intrinsically encode semantic correspondences between textual tokens and visual regions, ControlMLLM++ optimizes a latent visual token modifier during inference via a task-specific energy function to steer model attention towards user-specified areas. To enhance optimization stability and mitigate language prompt biases, ControlMLLM++ incorporates an improved optimization strategy (Optim++) and a prompt debiasing mechanism (PromptDebias). Supporting diverse visual prompt types including bounding boxes, masks, scribbles, and points, our method demonstrates strong out-of-domain generalization and interpretability. The code is available at \url{https://github.com/mrwu-mac/ControlMLLM}.
}

\keywords{Multimodal Large Language Models; Test-Time Computing; Referring MLLMs}



\maketitle

\section{Introduction}
Recent years have witnessed significant advances in large language models (LLMs) such as GPT-4~\citep{achiam2023gpt} and LLaMA~\citep{touvron2023Llama}, which demonstrate remarkable capabilities in understanding and generating human language. Inspired by these successes, considerable research effort has been devoted to extending LLMs toward multimodal understanding by integrating visual inputs, giving rise to multimodal large language models (MLLMs)~\citep{liu2024visual,li2023blip,dai2024instructblip,ye2023mplug,zhu2023minigpt,luo2024cheap}. Despite their impressive language-vision alignment, existing MLLMs primarily rely on coarse image-level correspondence and lack region-level understanding. This limits users’ ability to refer explicitly to specific image regions for detailed descriptions or reasoning, as text prompts alone often fail to adequately express intricate visual information.

To address this limitation, recent works have incorporated referring capabilities into MLLMs, enabling users to provide visual prompts such as bounding boxes, masks, or points to specify regions of interest~\citep{you2023ferret,bai2023qwen,zhang2023gpt4roi,chen2023shikra,lu2023lyrics}. While effective, these approaches generally require substantial training or fine-tuning on large annotated datasets containing region-text pairs, resulting in high computational costs and limited adaptability to new data domains or base models.

In contrast, we present \textbf{ControlMLLM++}, a test-time adaptation framework that endows pre-trained MLLMs with referring abilities \emph{without any model retraining or fine-tuning}. Our key insight is that the cross-modal attention mechanism inherently encodes rich semantic relationships between textual tokens and visual regions within MLLMs. By optimizing a learnable latent variable appended to the visual token embedding during inference, guided by a task-aware energy function on the aggregated attention maps, ControlMLLM++ effectively steers the model’s focus toward the user-specified region. This enables flexible and precise region-level control using diverse visual prompts—including bounding boxes, masks, scribbles, and points—while preserving the base MLLM’s language capabilities.

Moreover, ControlMLLM++ incorporates two core advancements that enhance the robustness and reliability of this test-time optimization process. First, an improved optimization strategy, \emph{Optim++}, accelerates convergence by selectively focusing on attention maps related to key text tokens and intermediate model layers. Second, a novel \emph{PromptDebias} mechanism mitigates overreliance on linguistic priors during inference, reducing prompt language bias and multimodal hallucination. Together, these components yield stable, accurate, and interpretable referring behaviors in diverse scenarios.

In summary, our contributions are threefold:

\begin{itemize}
    \item We introduce ControlMLLM++, a novel test-time latent variable optimization framework that injects explicit visual prompts into frozen pre-trained MLLMs to enable referring capabilities without additional training.
    \item We propose an enhanced optimization strategy, Optim++, and a PromptDebias mechanism to improve optimization stability and reduce language bias, resulting in more reliable and interpretable models.
    \item ControlMLLM++ effectively supports multiple forms of visual prompting (box, mask, scribble, point) and achieves strong out-of-domain generalization across various benchmarks, opening promising directions for controllable region-level visual reasoning in MLLMs.
\end{itemize}

\begin{figure*}[t]
  \centering
    \includegraphics[width=0.85\linewidth]{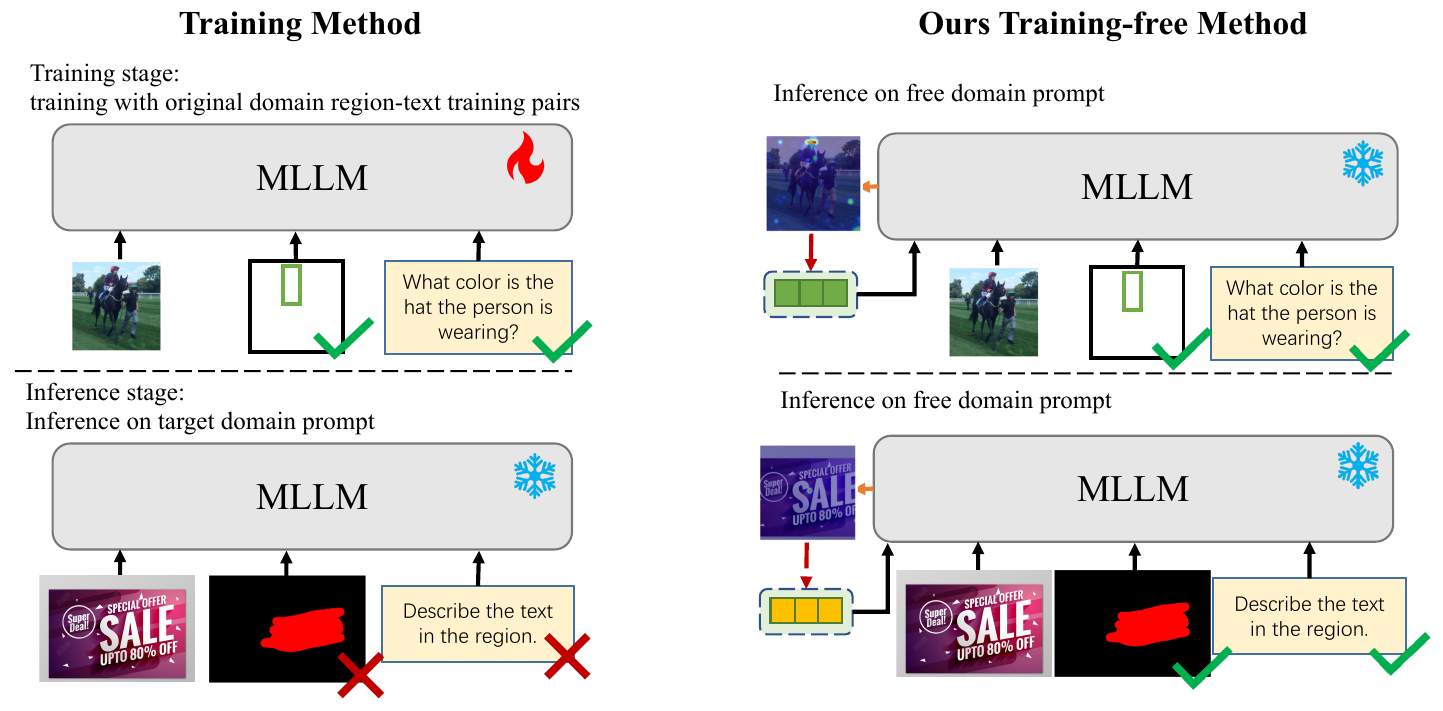}
  \caption{Comparison between the training method and our test-time computing method. The training method typically requires a large amount of in-domain data for training and cannot generalize to out-of-domain prompts. In contrast, our method can easily adapt to prompts from a new domain in a test-time computing manner.}
  \label{fig:intro}
\end{figure*}

\section{Related Work}
\label{sec:related}

\subsection{MLLMs}
Motivated by the accomplishments of Large Language Models (LLMs)~\citep{achiam2023gpt, touvron2023Llama}, there is a burgeoning trend among researchers to develop a diverse range of Multimodal Large Language Models (MLLMs)~\citep{li2023blip, liu2024visual, dai2024instructblip, ye2023mplug,li2023otter,luo2024cheap,luo2024feast,internlmxcomposer2_4khd, gao2023llamaadapterv2,zhang2023prompt,chen2023internvl,wang2023cogvlm,liu2023improved,fei2024vitron,fei2024enhancing,fei2024video,hu2024minicpm,zhang2025mllms,bai2025qwen2,zhu2025internvl3}. These MLLMs typically comprise a visual encoder, a language decoder, and an image-text alignment module~(vision-language connector. The visual encoder and the language decoder are often sourced from pre-trained models, such as CLIP~\citep{radford2021learning}, DINOv2~\citep{oquab2023dinov2}, Llama~\citep{touvron2023Llama}, and Vicuna~\citep{vicuna2023}. Meanwhile, the image-text alignment module is trained on image-text pairs and fine-tuned through visual instruction tuning to enhance its visual conversation capabilities. These Multimodal Large Language Models (MLLMs) often confront limitations stemming from their reliance on coarse image-level alignments.

\subsection{Referring MLLMs}
In recent research, there has been a noticeable trend towards integrating foundation models with tasks involving referring dialogue. These models~\citep{you2024ferret,zhang2024ferret,zhang2023gpt4roi,chen2023shikra,lu2023lyrics,peng2023kosmos,you2023ferret,xuan2023pink,yue2024sc,bai2023qwen,zhang2023llava,ma2024groma,he2024multi,lin2024draw,yuan2023osprey,zhao2023chatspot,chen2023position,sun2023alpha,rasheed2023glamm,zhou2023regionblip,xu2023pixel,cai2023making,guo2024regiongpt,ranasinghe2024learning,tian2024chatterbox,zhan2024griffon,heo2025omni,lim2025ureca} introduce spatial visual prompts as extra input and are trained using region-text pairs. By leveraging this approach, they effectively bridge the gap between textual prompts and visual context, enabling comprehensive understanding of image content at the regional level. 
While region-aware VLMs represent a significant leap in fine-grained vision-language understanding, their training complexity and resource demands remain critical bottlenecks. In contrast, our approach provides this capability to models without any training cost, achieving region-aware understanding through a novel test-time adaptation framework.

\subsection{Training-free Control in Text-to-Image}
There are numerous works on controllable text-to-image generation, among which training-free methods~\citep{hertz2022prompt,chen2024training,xie2023boxdiff,kim2023dense} are most relevant to our research. Among them, Prompt-To-Prompt~\citep{hertz2022prompt} explore the role of attention in text-visual interactions in Stable Diffusion~\citep{rombach2021highresolution} model, while Layout-Guidance~\citep{chen2024training} indirectly bias attention in Stable Diffusion model by optimizing an energy function. These contributions significantly inform our investigation into enhancing controllability and interpretability in MLLMs.

\subsection{Visual Prompt}
The visual prompt can be categorized into two main techniques: hard prompt and soft prompt.
The hard visual prompt works~\citep{shtedritski2023does,wang2023caption,yang2024fine,yang2023set} that direct the model's attention to the region or enable visual grounding abilities in the Multimodal Models in a training-free and convenient manner by directly manipulating images, such as color guidance~\citep{pmlr-v235-wu24l,gong2022person,yao2024cpt}. However, these methods inevitably compromise the structural information of the images, or a strong understanding of the corresponding patterns by the model is required. In contrast, the soft visual prompt works~\citep{jia2022visual,bahng2022exploring,zhang2024exploring} integrate learnable visual prompts into models to adapt them for different downstream tasks. However, these methods do not support region guidance and require fine-tuning the model with downstream data.
In contrast, we optimize a learnable latent variable to support referring MLLM in the test time, without any downstream training data required, TPT~\citep{shu2022test} is most related to our work. 


\begin{figure*}[t]
  \centering
    \includegraphics[width=0.9\linewidth]{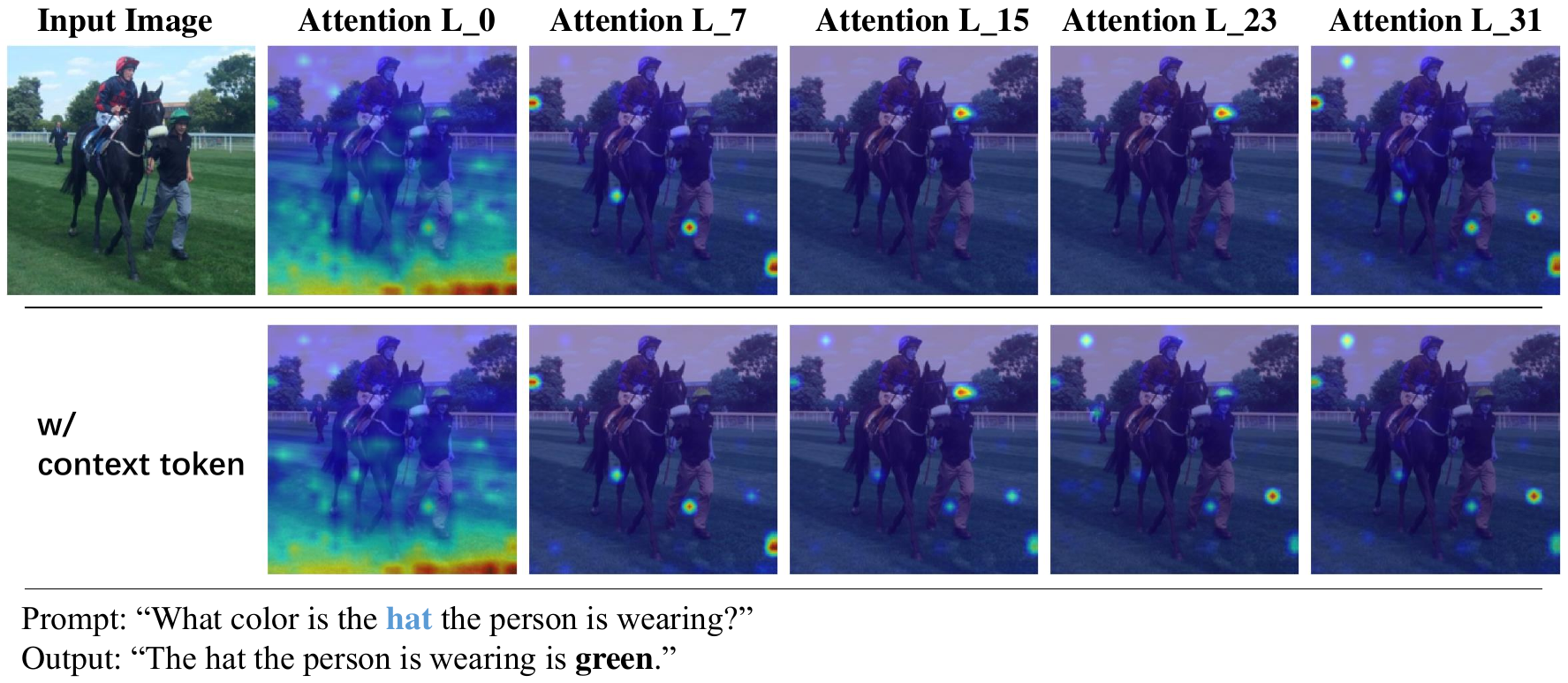}
  \caption{The attention maps in various layers of the MLLMs, with the numbers indicating the respective layer indices. The top line visualizes the attention between the prompt token ``hat'' and the visual tokens, while the bottom line visualizes the attention between the context token~(mentioned in Sec.~\ref{sec:ene}) and the visual tokens.}
  \label{fig:explain}
\end{figure*}

\section{Background}
\label{sec:background}

\subsection{Multimodal Large Language Models (MLLMs)}

Multimodal Large Language Models (MLLMs) integrate image and text inputs to perform joint understanding and generation. Typically, an MLLM consists of three components: a visual encoder, a vision-language connector, and a large language model decoder (LLM). Given an input image \(I\), the visual encoder outputs a set of visual tokens \(e_v = \{v_1, ..., v_{n_v}\}\), which are subsequently transformed by the vision-language connector into embeddings compatible with the LLM input space. Alongside textual tokens \(e_t = \{w_1, ..., w_{n_t}\}\) encoded from the text prompt \(p_t\), the concatenated embeddings \([e_v, e_t]\) serve as input to the LLM decoder. The LLM generates output tokens \(y = \{y_1, ..., y_{n_y}\}\) autoregressively as
\begin{equation}
\label{eq}
y_i = f_\theta(I, p_t, y_{<i}), \quad i=1,...,n_y,
\end{equation}
where \(y_{<i} = \{y_1, ..., y_{i-1}\}\) and \(\theta\) denotes the model parameters fixed during inference.

The core of the transformer-based decoder is the attention mechanism~\citep{vaswani2017attention}, especially the cross-attention layers where query tokens (text tokens or decoder hidden states) attend over key tokens (visual tokens). Formally, at attention layer \(\tau\), the cross-attention matrix is computed as
\begin{equation}
A^{(\tau)} = \text{softmax}\left(\frac{Q^{(\tau)} (K^{(\tau)})^\top}{\sqrt{d_k}}\right) \in \mathbb{R}^{n_q \times n_v},
\end{equation}
where \(n_q\) and \(n_v\) are the numbers of query and visual key tokens, respectively. Each entry \(A^{(\tau)}_{ij}\) indicates how strongly the \(i\)-th text query token attends to the \(j\)-th visual token in layer \(\tau\).

\subsection{Referring in MLLMs}

Referring MLLMs aim to extend an MLLM’s output conditioning to incorporate visual referring prompts \(r\), such as bounding boxes, masks, scribbles, or points, that indicate user-specified image regions. The goal is to produce answers grounded on both the text prompt \(p_t\) and the region \(r\):
\begin{equation}
y_i = f_\theta(I, p_t, r, y_{<i}),
\end{equation}
where \(r\) guides fine-grained spatial focus.

\paragraph{Training-based Approaches}

Most prior works achieve this by fine-tuning \(\theta\) on annotated datasets with region-text supervision~\citep{you2023ferret,bai2023qwen}. However, this requires costly retraining and often results in models specialized to training domains, limiting out-of-distribution robustness~\citep{shu2022test}.

\paragraph{Training-Free Referring}

A more flexible solution is to keep \(\theta\) frozen and embed referring prompts into the model input or internal representations at test time, enabling referring abilities without retraining. ControlMLLM++ falls into this category, performing test-time optimization of latent perturbations to visual tokens to steer attention maps towards the referred region \(r\).

\section{Method}
\label{sec:method}
We aim to propose a test-time computing method to overcome the inconveniences of traditional training. The referring MLLM via test-time computing maintains the model parameters \(\theta\) frozen, eliminating the need for any training or fine-tuning with samples from the training set. During inference, the only information available is the single test sample without label information, as shown in Figure~\ref{fig:intro}~(right).

\begin{figure*}[t]
  \centering
    \includegraphics[width=0.9\linewidth]{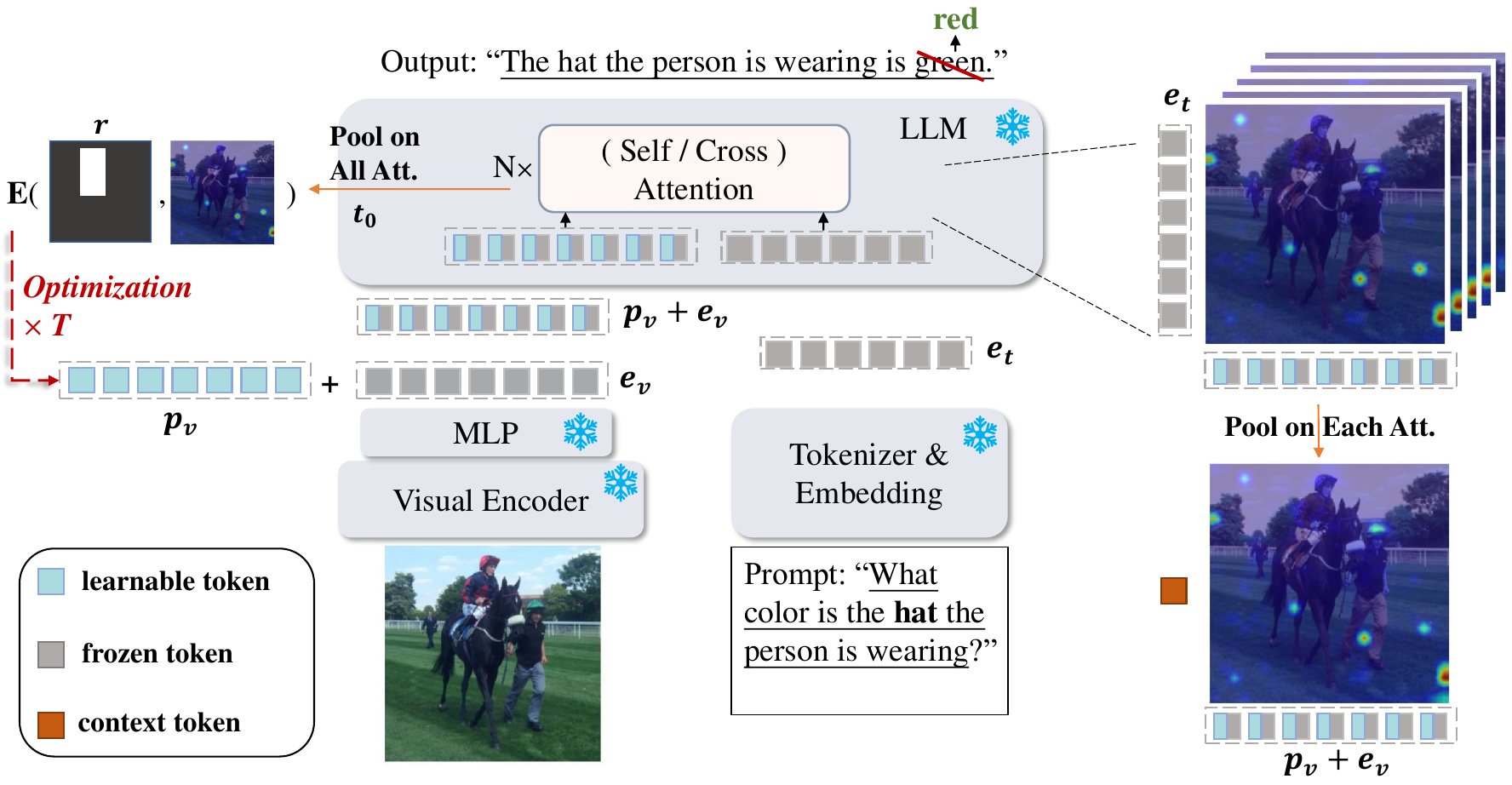}
  \caption{The overview of ControlMLLM. With the provided visual prompt, we convert it into a mask, and compute the mask-based energy function between the mask and the pooled attention map. During the inference process, we conduct backpropagation to optimize a learnable latent variable. This process is executed at the 0-th step of model inference and iterated $T$ times.}
  \label{fig:method}
\end{figure*}

In this section, we explore and design a solution to address the challenges of Referring MLLMs via test-time computing. The key task is to flexibly embed visual prompts during the inference phase while maintaining the model's reasoning capabilities. To begin with, we delve into the mechanism of MLLMs (see Sec.~\ref{sec:ana}), our key observation is the attention mechanism in LLM capturing the relationship between the model's output and the input pixels. Further, the visual tokens inputted into the LLM influence the values of the attention maps to indirectly control the model output. Building on this analysis, we propose the Latent Variable learning~(a test-time prompt tuning strategy~\citep{shu2022test}, see Sec.~\ref{sec:ene}) to edit the visual tokens, as shown in Figure~\ref{fig:method}. This method effectively integrates visual prompts into pre-trained MLLMs, enabling fine-grained visual reasoning.

\begin{figure*}[t]
  \centering
    \includegraphics[width=0.95\linewidth]{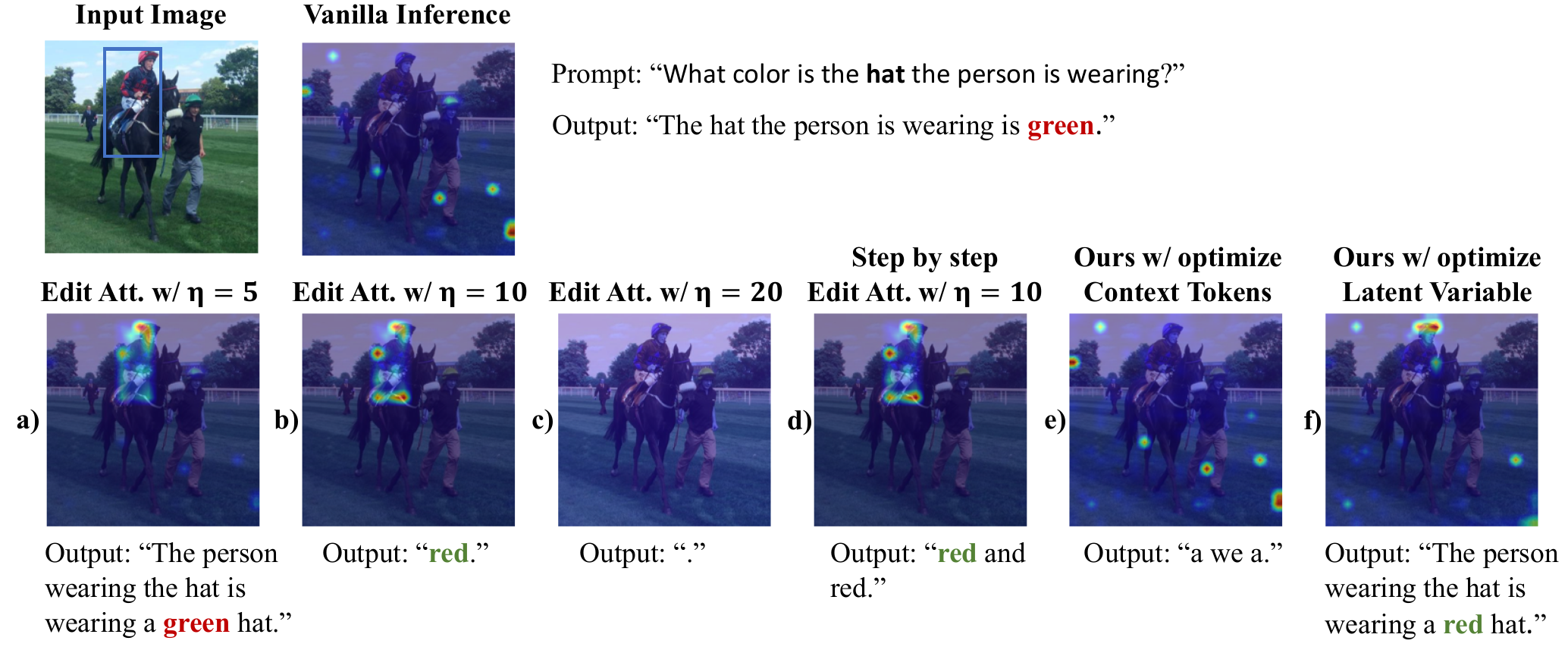}
  \caption{Manipulating attention with various methods: (a), (b), and (c) demonstrate the manipulation of the attention map by adding an adjustment coefficient $\eta$ on the attention map in the first step during the model inference. (d) illustrates the step-by-step editing approach. (e) showcases that optimizing a learnable context tokens~(mentioned in Sec.~\ref{sec:ene}) instead of visual tokens, while (f) presents the results of our method optimizing the learnable latent variable. }
  \label{fig:rel}
\end{figure*}

\subsection{Analysis of the Attention in LVLMs}
\label{sec:ana}
We begin by analyzing \textbf{\textit{which factors in the model truly capture the relationship between input and output?}} In other words, we seek to understand \textbf{\textit{how to interpret the association between the model's output and the input pixels}}. 

As demonstrated by Equation~\ref{eq1}, Multimodal Large Language Models (MLLMs) fundamentally model the maximum likelihood output based on visual input and text prompts. By conditioning on the text prompt, the model can determine which parts of the image have the greatest impact on the output. Building on the discussions in the Sec.~\ref{sec:background} and illustrations in the Figure~\ref{fig:explain}~(top line), we can observe that the attention map models the influence of visual tokens on the output conditioned by the text prompt. Therefore, the attention map in MLLMs not only provides interpretability regarding the relationship between model output and input pixels but also facilitates guiding the model's output.

A natural idea is that we can directly alter the model's output by editing the attention maps. Inspired by IBD~\citep{zhu2024ibd}, we achieve this by adding an adjustment coefficient $\eta$ to the attention related to the visual tokens corresponding to the referring region, which can be formulated as:
\begin{equation}
\begin{aligned}
\label{eq:editatt}
& A^{(\tau)}=\text{softmax}(\frac{[e_v,e_t]^{(\tau)} \cdot ([e_v,e_t]^{(\tau)})^T}{\sqrt{d_k}} + M), \\
& M_i =\eta \quad \text { if } i \text{ in } r \quad \text { else } 0,
\end{aligned}
\end{equation}
where $M$ is a mask with the same shape as the attention map, $r$ denotes the referring region.
However, we have to carefully select a suitable coefficient $\eta$ for each example. When the $\eta$ is too small, it leads to ineffective control (as shown in Figure~\ref{fig:rel} a), and when it is too large, it can impact the language capabilities of the LLM (as shown in Figure~\ref{fig:rel} c). Additionally, we found that it is sufficient to manipulate the attention map at the 0-th step during model inference (as shown in Figure~\ref{fig:rel} a,b,c), as it is most directly associated with the text prompt, and manipulating attentions step by step also affects the expression of the LLM (as shown in Figure~\ref{fig:rel} d). Overall, directly manipulating attention maps is not a viable approach because it overlooks the relationships between attention layers and not all layers' visual tokens decide the output~\citep{chen2024image,wu2024accelerating}.

We note that in MLLMs, typically a vision-language connector is trained for image-text alignment. This implies that MLLMs indirectly affect the values of the attention map by learning the parameters of the connector to alter the visual tokens. In other words, the visual tokens inputted into the LLM directly influence the values of the attention maps.


\subsection{ControlMLLM: Manipulating Attention via Latent Variable Learning}
\label{sec:ene}
Based on the analysis above, our core idea is to indirectly influence the attention maps by editing visual tokens, thereby focusing on the referred regions. We achieve this by optimizing a learnable latent variable based on an energy function~\citep{chen2024training,wu2024tradiffusion}, which calculates the relationship between the input referring and the attention maps. To do this, we first need to determine which attention maps to use. One approach is to use attention maps between each text prompt token and all visual tokens. However, because visual tokens typically have a significant impact on the result based on only a few most relevant text prompts (referred to as \textbf{highlight text tokens}), using all attention maps would be computationally redundant. Yet, for users, identifying the highlight text tokens can be challenging. Therefore, we simply average pool the attention maps generated for each text prompt token to represent the global context of the text prompt (referred to as the \textbf{context token}) and its association with visual tokens. We found that this simple method of using context tokens produces attention maps similar to those generated by highlight text tokens, as shown in Figure~\ref{fig:explain}~(bottom line). We leave the optimization based on highlight text tokens for future work.

Specifically, our method supports four types of referring shapes, including box, mask, scribble, and point. We employ two types of energy functions to respectively support these referring shapes: a hard mask-based energy function for box and mask referring, and a soft mask-based energy function for scribble and point referring.

\subsubsection{Hard Mask-based Energy Function}
We first zero initialize a learnable latent variable $p_v$ with the same shape as $e_v$, and add it to the $e_v$. Then we can get $N$ attention maps from $N$ attention layers which model the relation between the context token and the novel visual tokens. Given the referring box or mask, we first convert it into a binary mask. Then, we compute the mask-based energy function based on the mask and the attention map $A^{(ct)}$, which is obtained by averaging pooling from $N$ attention maps. The energy function can be formulated as:
\begin{equation}
E\left(A^{(ct)}, r\right)=\left(1-\frac{\sum_{i \in r} A_{i}^{(ct)}}{\sum_i A_{i}^{(ct)}}\right)^2,
\label{eq2}
\end{equation}
where $r$ denotes the referring region. Then the gradient of the loss~\ref{eq2} is computed via backpropagation to update the learnable latent variable:
\begin{equation}
\boldsymbol{p}_v \leftarrow \boldsymbol{p}_v- \alpha \nabla_{\boldsymbol{p}_v} E\left(A^{(ct)}, r\right),
\label{eq3}
\end{equation}
where $\alpha>0$ is a hyperparameter controlling the strength of
 the guidance. By optimizing $p_v$ through the Equation~\ref{eq3}, we indirectly guide the attention maps to produce higher responses in the referring region $r$, thereby increasing the influence of the visual content of region $r$ on the output.

\subsubsection{Soft Mask-based Energy Function}
Since scribble and point lack the concept of the region, 
it is optional to use an extra SAM~\citep{kirillov2023segment} model to obtain a mask for applying the Hard Mask-based Energy Function. However, this incurs additional inference cost, so we also provide an optional soft mask-based energy function based on a distance matrix $D$, which is computed via applying the OpenCV~\citep{bradski2000opencv} \textit{distanceTransform} function on the given scribble or point. Then the soft mask-based energy function can be formulated as: 
\begin{equation}
E\left(A^{(ct)}, r\right)=\left(1-\frac{\sum_{i \in r} \frac{{e^{-{{D_i^2}}/{{2 \sigma^2}}}}}{{\sqrt{2 \pi} \sigma}}
 A_{i}^{(ct)}}{\sum_i A_{i}^{(ct)}}\right)^2,
\label{eq4}
\end{equation}
where $\sigma$ is the standard deviation of the Gaussian function, which is set to 0.1. By optimizing $p_v$ through the Equation~\ref{eq4}, the closer the region of attention map is to the given scribble or point, the higher the response.

Finally, we iteratively optimize the learnable latent variable $T$ times at the $0$-th step of model inference. 
Since ControlMLLM employs an overly simplistic optimization strategy, we adopt Early Stopping (ES) and Exponential Moving Average (EMA) to prevent overfitting and stabilize the optimization process.
More details are shown in Appendix.

\subsection{ControlMLLM++: Enhanced ControlMLLM with Enhanced Optimization and Language Bias Mitigation}
Although Early Stopping and Exponential Moving Average (EMA) strategies help stabilize the optimization process, achieving optimal performance requires careful hyperparameter tuning and often results in slow convergence. To address this, we conducted an in-depth analysis of the impact of attention layers, text tokens, and prompt language biases on optimization outcomes. Based on our findings, we propose an enhanced optimization metric, Optim++, and a language bias mitigation metric, PromptDebias, to further improve performance and robustness.

\begin{figure}[t]
  \centering
\includegraphics[width=0.99\linewidth]{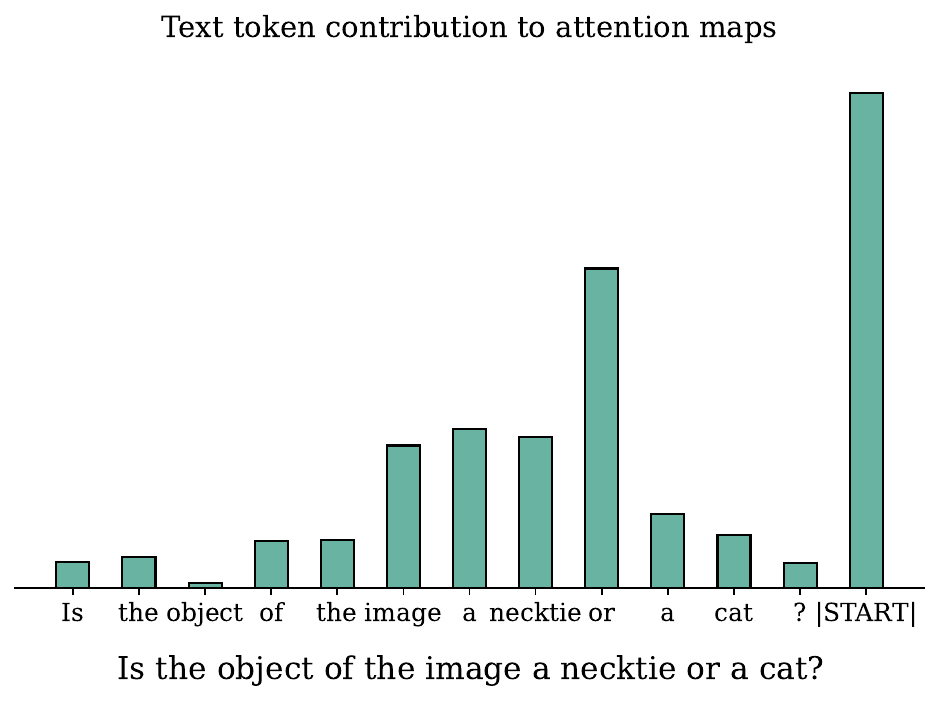}
 \caption{The text tokens' contribution to attention maps, where the attention is focused on the answer-start token~($|$START$|$).}
    \label{fig:optim-a}
\end{figure}

\begin{figure}[t]
  \centering
\includegraphics[width=0.99\linewidth]{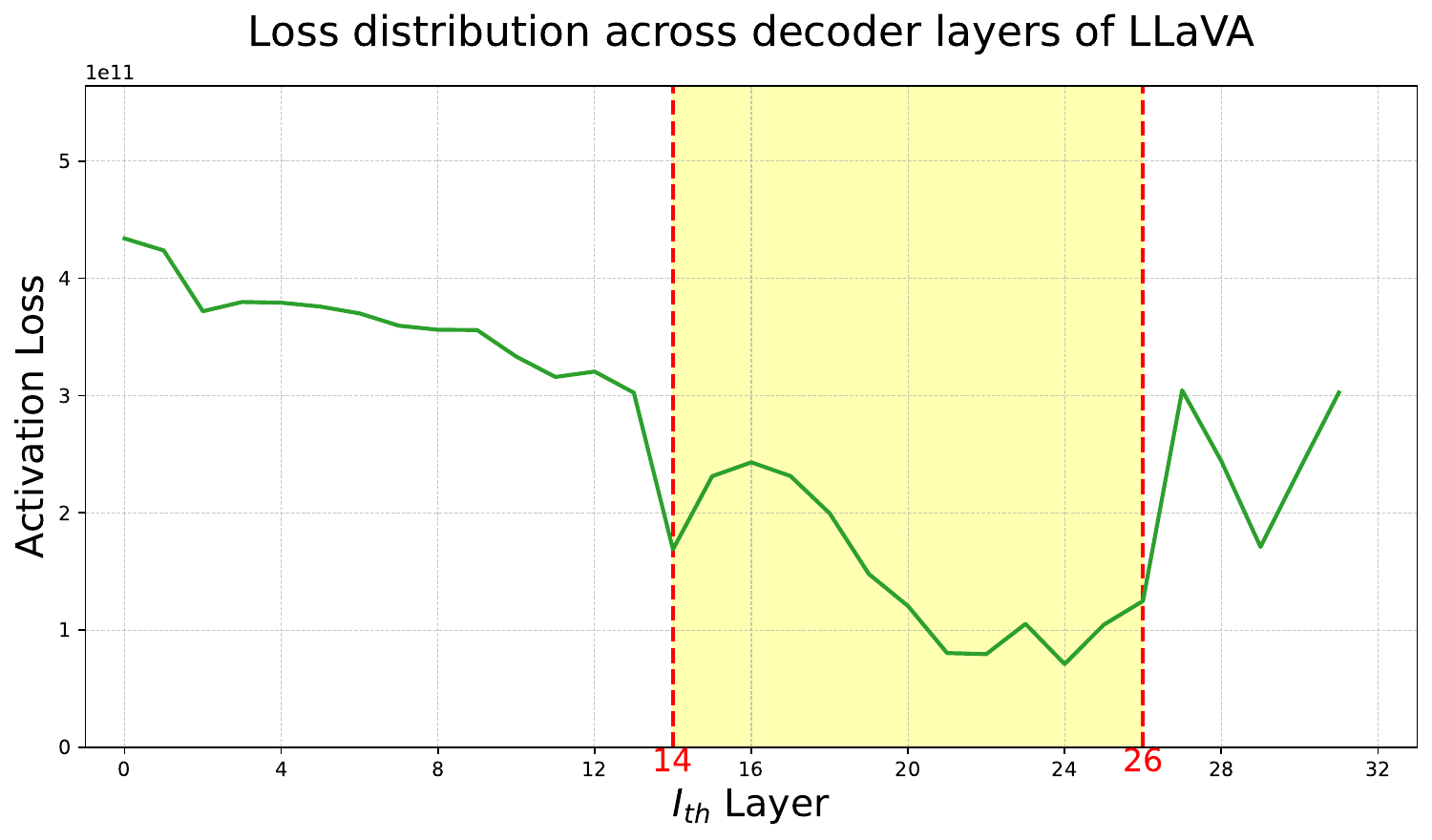}
  \caption{The loss distribution across decoder layers, showing that the text-visual attention is primarily concentrated in the middle layers.}
    \label{fig:optim-b}
\end{figure}

\begin{figure}[t]
  \centering
\includegraphics[width=0.99\linewidth]{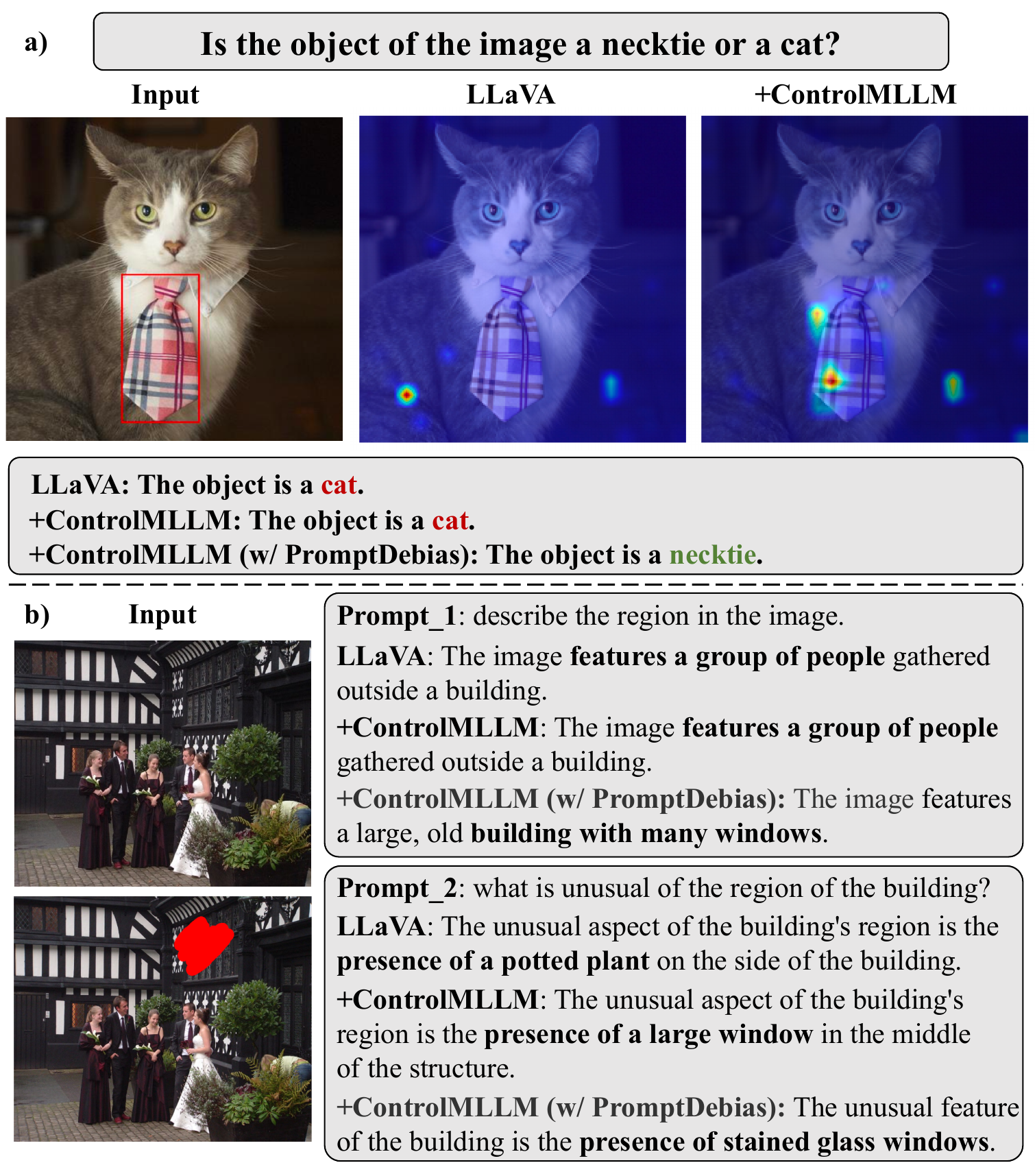}
  \caption{a) An example where ControlMLLM focus on the correct region, yet produce the same unexpected output as LLaVA due. b) Different prompts lead to different optimization performance. Such behavior is likely caused by the model's overreliance on linguistic priors.}
  \label{fig:debias}
\end{figure}

\begin{figure*}[t]
  \centering
    \includegraphics[width=0.9\linewidth]{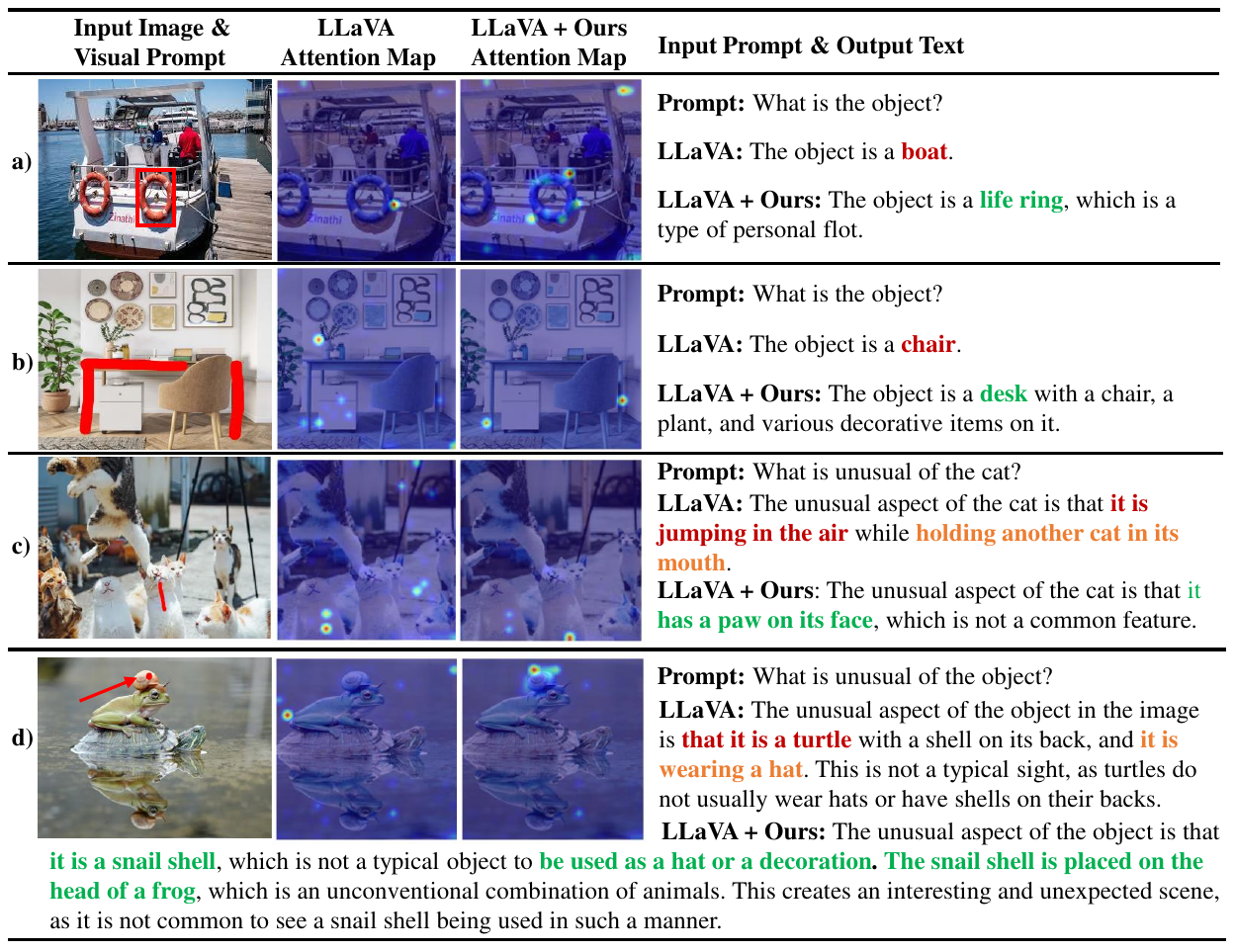}
  \caption{The examples of referring MLLM with four types of visual prompt, including box~(a), mask~(b), scribble~(c) and point~(d). The correct referring expressions are marked in green, incorrect referring expressions are marked in red, and hallucinated expressions are marked in orange. Compared to the baseline model, our method enhances \textbf{interpretability} and \textbf{controllability} with visual prompts, while also helping the model \textbf{mitigate hallucination} issues.}
  \label{fig:vis1}
\end{figure*}

\begin{figure*}[t]
  \centering
    \includegraphics[width=0.9\textwidth]{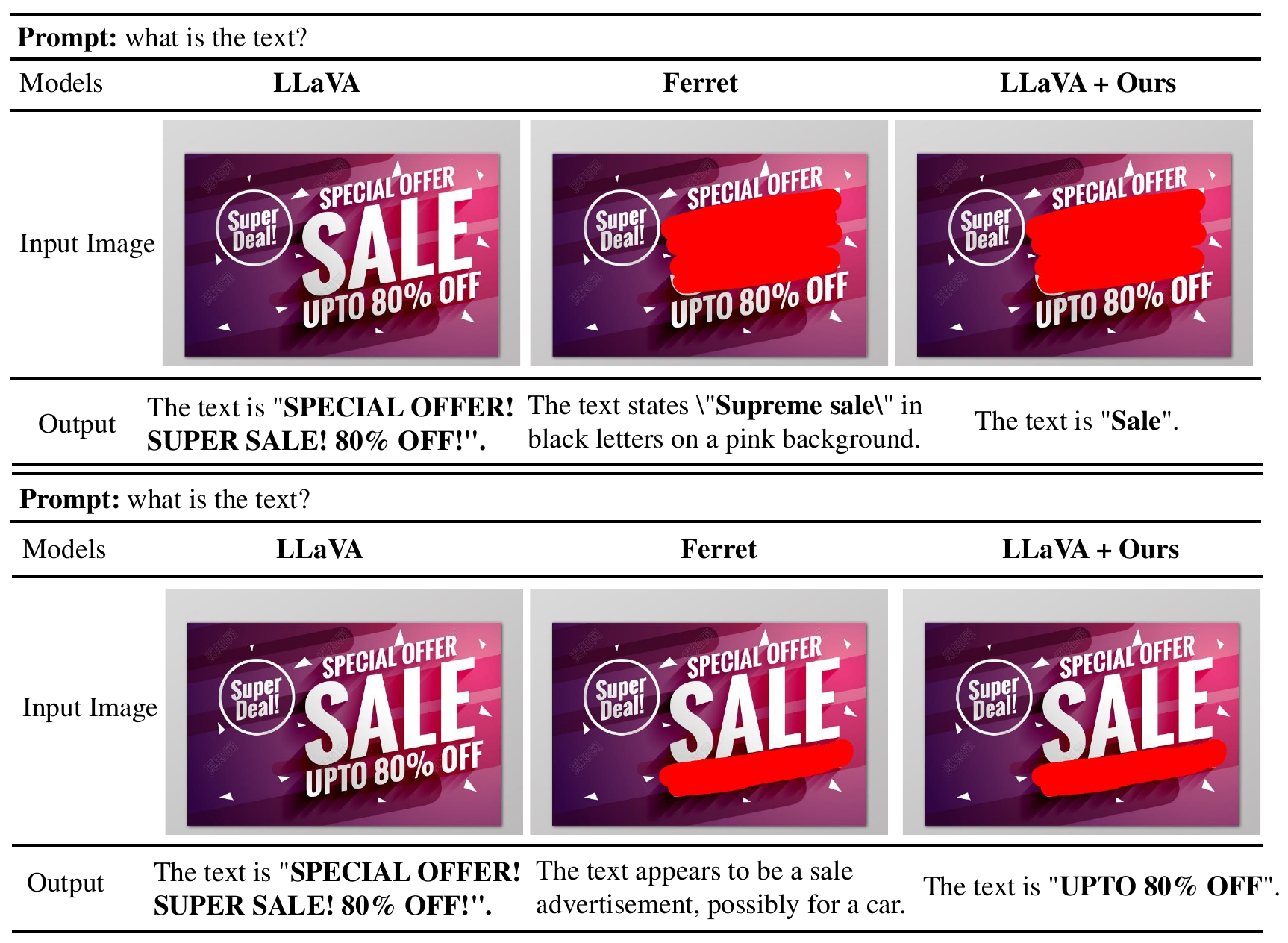}
  \caption{Comparison of OCR performance across models. Ferret, though trained with grounding supervision, still struggles with localized understanding on OCR task. In contrast, our method (LLaVA + Ours) accurately identifies the referred textual content.}
  \label{supfig:com_ocr}
\end{figure*}

\subsubsection{Enhanced Optimization with Optim++}
In ControlMLLM, we initially optimize the attention maps between context tokens and visual tokens across all layers. However, this approach includes many irrelevant tokens and attention layers that are not crucial to the optimization process, leading to slower convergence. Ideally, we hope the model to attend to the user-specified region when generating its response. Since ControlMLLM performs latent optimization at the first decoding step, this implies that the answer-start token should be encouraged to focus on the referred region from the very beginning. As shown in Figure~\ref{fig:optim-a}, the attention is focused on the answer-start token. Additionally, as depicted in Figure~\ref{fig:optim-b}, we observe that the loss in layers 14-26 of the LLaVA decoder is significantly lower than in other layers, indicating that the attention maps capturing text-visual relationships are primarily concentrated in these middle layers.

Based on these findings, we adjust the original attention map $A^{(ct)}$ used in ControlMLLM for energy function calculation by applying average pooling to the attention maps between the answer-start token and visual tokens within these middle layers, as $A^{(st)}$. This strategy significantly reduces the complexity of optimization, improving both inference efficiency and optimization performance. To further enhance optimization speed and stability, we replace the previous Gradient Descent, EMA, and Early Stopping strategies with the Adam optimizer, which requires less meticulous tuning. The updated optimization process is as follows:
\begin{align}
\boldsymbol{m}_t &= \beta_1 \boldsymbol{m}_{t-1} + (1 - \beta_1) \nabla_{\boldsymbol{p}_v} \alpha \cdot E\left(A^{(st)}, r\right), \\
\boldsymbol{v}_t &= \beta_2 \boldsymbol{v}_{t-1} + (1 - \beta_2) \left(\nabla_{\boldsymbol{p}_v} \alpha \cdot E\left(A^{(st)}, r\right)\right)^2, \\
\hat{\boldsymbol{m}}_t &= \frac{\boldsymbol{m}_t}{1 - \beta_1^t}, \quad \hat{\boldsymbol{v}}_t = \frac{\boldsymbol{v}_t}{1 - \beta_2^t}, \\
\boldsymbol{p}_v &\leftarrow \boldsymbol{p}_v - lr \cdot \frac{\hat{\boldsymbol{m}}_t}{\sqrt{\hat{\boldsymbol{v}}_t} + \epsilon},
\end{align}
where \( \boldsymbol{m}_t \) represents the first moment (mean) of the gradient, while \( \boldsymbol{v}_t \) denotes the second moment (uncentered variance) of the gradient. The terms \( \beta_1 \) and \( \beta_2 \) are the decay rates for \( \boldsymbol{m}_t \) and \( \boldsymbol{v}_t \), respectively. The learning rate is denoted by \( lr \), and \( \epsilon \) is a small constant added to avoid division by zero during updates. The parameters \( \hat{\boldsymbol{m}}_t \) and \( \hat{\boldsymbol{v}}_t \) are bias-corrected versions of \( \boldsymbol{m}_t \) and \( \boldsymbol{v}_t \), adjusted to account for initialization bias.

\subsubsection{Language Bias Mitigation with PromptDebias}
During the optimization process, we also observe a language bias issue in the model. As shown in the Figure~\ref{fig:debias}~(a), where ControlMLLM focus on the correct region, yet produce the same unexpected output as LLaVA. And even prompts with similar semantics can significantly impact the optimization performance of ControlMLLM as shown in Figure~\ref{fig:debias}~(b). This suggests that the model tends to favor generating certain types of outputs depending on the specific phrasing of the prompt. Such behavior is likely caused by the model's overreliance on linguistic priors~\citep{wu2022difnet,zhang2021rstnet} while overlooking visual information—a manifestation of multimodal hallucination~\citep{leng2024mitigating,pmlr-v235-wu24l}.

To address this issue, inspired by VCD~\citep{leng2024mitigating}, we propose PromptDebias, a contrastive decoding strategy designed to mitigate the model's prompt bias. Specifically, during the optimization process, we obtain two types of conditional outputs: one generated without the visual prompt and the other with it. By combining the output logits based on both conditions, we encourage the model to produce less biased results. 
Specifically, we compute the final token probability distribution as:
\begin{equation}
\begin{aligned}
    p(y) = \text{softmax}(& (1 + \gamma) \cdot \text{logit}_\theta(y \mid I, p_t, p_v) \\
                          & - \gamma \cdot \text{logit}_\theta(y \mid I, p_t)),
\end{aligned}
\end{equation}
where $\text{logit}_\theta(y \mid I, p_t, p_v)$ denotes the logit predicted by the model given the image $I$, textual prompt $p_t$, and visual prompt $p_v$, while $\text{logit}_\theta(y \mid I, p_t)$ refers to the logit obtained without visual prompt input. The scalar hyperparameter $\gamma$ controls the strength of the contrastive signal.

This approach effectively reduces the model's overdependence on linguistic priors. As a result, the model is encouraged to attend more to the injected visual cues during decoding.

\section{Experiments}
\subsection{Experiment Details}
\label{sec:detail}
Unless explicitly stated otherwise, the MLLM we use is LLaVA-v1.5-7B~\citep{liu2023improved}. For ControlMLLM, we set $T$=5, $\alpha$=400 and $\beta=0.5$. For ControlMLLM++, we set $T$=3, $\alpha$=400, $lr$=0.03 and $\gamma=0.7$. All experiments are conducted on two RTX 4090 GPUs with 24 GB of memory each. 

\subsection{Applications}
\subsubsection{Referring with Different Visual Prompts}
We first demonstrate referring QA with different visual prompts, including box, mask, scribble and point in the Figure~\ref{fig:vis1}.  Our method consistently demonstrates significant controllability with four types of visual prompts. And our method improves the interpretability compared to basic model~(column 3 vs column 2), demonstrates a stronger correlation between the attention response areas and the generated descriptions.

\subsubsection{Out-of-Domain Task}
We present examples of the performance on out-of-domain tasks OCR. As shown in Figure~\ref{supfig:com_ocr}, compared to Ferret, our method correctly identified the text in the referring region. 

\subsubsection{Impact on Hallucinations} 
Our method guides the model to focus on specific regions, potentially helps the model mitigate hallucination issues, as shown in Figure~\ref{fig:vis1}~(c,d output in orange color).

\subsubsection{Comparison on Referring Object Classification Task}
Following Ferret~\citep{you2023ferret,zhang2024ferret}, we use the Referring Object Classification~(ROC) task to evaluate whether our method can accurately pinpoint and understand the semantic of the referring region. The task requires the model to correctly identify the target within the referring region. We follow the setting of Ferret to form 1,748 questions~(in which 1,548 for test and 200 for validation) based on LVIS~\citep{gupta2019lvis} validation dataset, with corresponding box, mask, scribble and point. 
We consider the edit attention with $\eta=10$~(as Equation~\ref{eq:editatt} and Figure~\ref{fig:rel}~(b)) as the baseline model. And we compare several training methods~\citep{peng2023kosmos,zhang2023gpt4roi,chen2023shikra,you2023ferret}. 
We also evaluate the lower and upper limits of LLaVA's recognition capability by assessing LLaVA without referring region, as well as background blur outside the referring region, which are presented in gray. Additionally, we evaluate a method that highlights regions with color as a comparable training-free method.
More details and the input examples are shown in Appendix.


The results are shown in Table~\ref{Tab1}. ControlMLLM shows a better performance than the training method GPT4-ROI with box referring~(60.59 vs 58.59) and the Shikra-7B with point referring~(58.85 vs 56.27). And ControlMLLM++ presents a similar performance compared to the training method Ferret~\citep{you2023ferret}. Our method also demonstrates superiority compared to training-free color prompt-based method and baseline method. 

\subsubsection{Comparison on Referring Text Classification Task}
We consider the Referring Text Classification~(RTC) task as the \textbf{out-of-domain} task, to verify the model's out-of-domain transfer capability. Similar to the ROC task, we formulate the problem as a binary classification task and construct 1,372 questions~(in which 1,172 for test and 200 for validation) based on the COCO-Text~\citep{veit2016coco} dataset. Since point and scribble referring methods are not suitable for text due to the non-connectivity of the text, we only evaluate the RTC task with box and mask referring. 
Since the RTC task converges faster, we set T=4 for ControlMLLM and T=1 for ControlMLLM++.

The results are shown in Table~\ref{Tab1}. All the training methods we evaluated exhibited poor out-of-domain generalization performance. Specifically, Ferret achieves only 58.28\% accuracy on the RTC task, despite its excellent in-domain performance as shown in Table~\ref{Tab1}. In contrast, our test-time method still demonstrates the best \textbf{out-of-domain generalization} performance. We also present comparative examples of out-of-domain tasks, as shown in Figure~\ref{supfig:com_ocr} and Figure~\ref{supfig:com_screen}.

\begin{table*}[t]
\caption{Performance on Referring Object Classification (ROC) and Referring Text Classification (RTC) tasks. ROC prompts follow the template: “\textit{Is the object \{location\} a \{class A\} or a \{class B\}?}”; RTC prompts: “\textit{Is the text \{location\} ‘\{text A\}’ or ‘\{text B\}’? Please select only one.}” “--” indicates unsupported input types. Gray numbers are for reference only.}

\label{Tab1}
\begin{small}
\begin{sc}
\begin{tabular}{l|cccc|cc}
    \toprule
    \multirow{2}{*}{Models} & \multicolumn{4}{c}{ROC~(In-Domain)} & \multicolumn{2}{c}{RTC~(Out-of-Domain)} \\
    \cmidrule(lr){2-5} \cmidrule(lr){6-7}
     & Box & Mask & Scribble & Point & Box & Mask \\
    \midrule
    \small\textit{Training Methods:} \\
    Kosmos-2~\citep{peng2023kosmos} & 55.17 & - & - & - & 23.12 & - \\
    GPT4-ROI~\citep{zhang2023gpt4roi} & 58.59 & - & - & - & 57.17 & - \\
    Shikra-7B~\citep{chen2023shikra} & 64.60 & - & - & 56.27 & 60.41 & - \\
    Ferret-7B~\citep{you2023ferret} & 71.71 & 72.39 & 71.58 & 68.54 & 58.70 & 59.81 \\
     
    \midrule
    \small\textit{Training-Free Methods:} \\
    \textcolor{gray}{LLaVA-1.5}~\citep{liu2024visual} & \textcolor{gray}{54.72} & \textcolor{gray}{54.72} & \textcolor{gray}{54.72} & \textcolor{gray}{54.72} & \textcolor{gray}{57.42} & \textcolor{gray}{57.42} \\
    \textcolor{gray}{LLaVA-1.5 + Blur} & \textcolor{gray}{73.39} & \textcolor{gray}{71.32} & - & - & \textcolor{gray}{84.62} & \textcolor{gray}{75.85} \\
    LLaVA-1.5 + Color & 55.10 & 56.72 & - & - & 59.04 & 56.57 \\
    LLaVA-1.5 + Edit Att & 36.24 & 37.08 & - & - & 50.00 & 51.88 \\
    LLaVA-1.5 + \textbf{ControlMLLM} & 60.59 & 60.79 & 58.33 & 58.85 & 63.06 & 61.43 \\
    LLaVA-1.5 + \textbf{ControlMLLM++} & \textbf{71.19} & \textbf{73.00} & \textbf{65.12} & \textbf{65.13} & \textbf{74.66} & \textbf{74.65} \\
    \bottomrule
\end{tabular}
\end{sc}
\end{small}
\end{table*}

\begin{table}[t]
\caption{The results of combining with different MLLMs on ROC and RTC tasks~(box, test set).}
\label{supTab1}
\setlength{\tabcolsep}{4pt}
\begin{small}
\begin{sc}
\begin{tabular}{lcc}
    \toprule
     Models & ROC & RTC  \\
    \midrule
     \textcolor{gray}{LLaVA-1.5}~\citep{liu2024visual} & \textcolor{gray}{54.72} & \textcolor{gray}{57.42} \\
     LLaVA-1.5 + ControlMLLM & 
    60.59 & 63.06 \\
    LLaVA-1.5 + ControlMLLM++ & \textbf{71.19} & \textbf{74.66} \\
    \midrule
     \textcolor{gray}{LLaVA-HR}~\citep{luo2024feast} & \textcolor{gray}{53.81} & \textcolor{gray}{57.00} \\		
     LLaVA-HR + ControlMLLM &
    58.92 & 66.89 \\ 
    LLaVA-HR + ControlMLLM++ & \textbf{69.06} & \textbf{82.68} \\
    \midrule
    Qwen2.5-VL~\citep{bai2025qwen2} & 78.81 & 81.91\\
    Qwen2.5-VL + ControlMLLM & 79.20 & 86.43 \\
    Qwen2.5-VL + ControlMLLM++ & 79.20 & \textbf{88.23} \\
    \bottomrule
    \end{tabular}
\end{sc}
\end{small}
\end{table}


\subsubsection{More Tasks and MLLMs}
we validate our method on the more MLLMs, including LLaVA-1.5, Qwen2.5-VL-7B~\citep{bai2025qwen2} and LLaVA-HR-7B~\citep{luo2024feast}. LLaVA-1.5 and LLaVA-HR, which lack native referring capabilities. In contrast, Qwen2.5-VL, a recently released MLLM equipped with built-in referring functionality.
We observe that ControlMLLM++ brings consistent and substantial improvements across all models, regardless of whether they have built-in referring capabilities. For LLaVA-1.5 and LLaVA-HR, our method significantly enhances performance, demonstrating its effectiveness in equipping base MLLMs with strong region understanding. Similarly, LLaVA-HR shows even greater gains on RTC, possibly due to higher resolution inputs benefiting text interpretation. Even for Qwen2.5-VL, a state-of-the-art MLLM already equipped with referring capabilities, ControlMLLM++ yields further improvement, especially on the RTC task. This suggests that our method is not only effective as a plug-in module for grounding-agnostic models but can also enhance the robustness and domain generalization of referring-capable MLLMs. These results validate that ControlMLLM++ is a generalizable and effective test-time plug-in that enhances referring understanding across a wide spectrum of MLLM architectures.

Table~\ref{tab:pami_style_rd} presents the results of our method on the box-level referring description task over two datasets: RefCOCOg~\citep{kazemzadeh2014referitgame}~(in-domain) and Screenshot~\citep{cheng2024seeclick}~(out-of-domain). We report four standard language generation metrics: BLEU@4~(B@4)~\citep{papineni2002bleu}, METEOR~(M)~\citep{banerjee2005meteor}, CIDEr-D~(C)~\citep{vedantam2015cider}, and SPICE~(S)~\citep{anderson2016spice}. Applying our training-free method, ControlMLLM, consistently improves the referring quality of all models. On RefCOCOg, it notably boosts LLaVA-HR’s CIDEr score from 56.29 to 68.82. The enhanced variant, ControlMLLM++, brings further gains across all models. For instance, LLaVA-HR + ControlMLLM++ achieves 78.42 CIDEr on RefCOCOg, outperforming Qwen2.5-VL, despite the latter’s built-in referring capability. 

When evaluated on the Screenshot dataset (out-of-domain), the performance gap between baselines becomes more pronounced. LLaVA and LLaVA-HR struggle due to their lack of fine-grained spatial understanding. Interestingly, even Qwen2.5-VL, although trained for referring, shows moderate poor in out-of-domain generalization. In this scenario, ControlMLLM++ exhibits strong robustness, confirming its adaptability across domains. These results highlight that our method not only equips non-referring models with effective grounding ability but also complements modern referring-capable MLLMs by enhancing their generalization and precision—particularly in zero-shot, domain-shifted scenarios.

\begin{figure*}[t]
  \centering
    \includegraphics[width=0.99\textwidth]{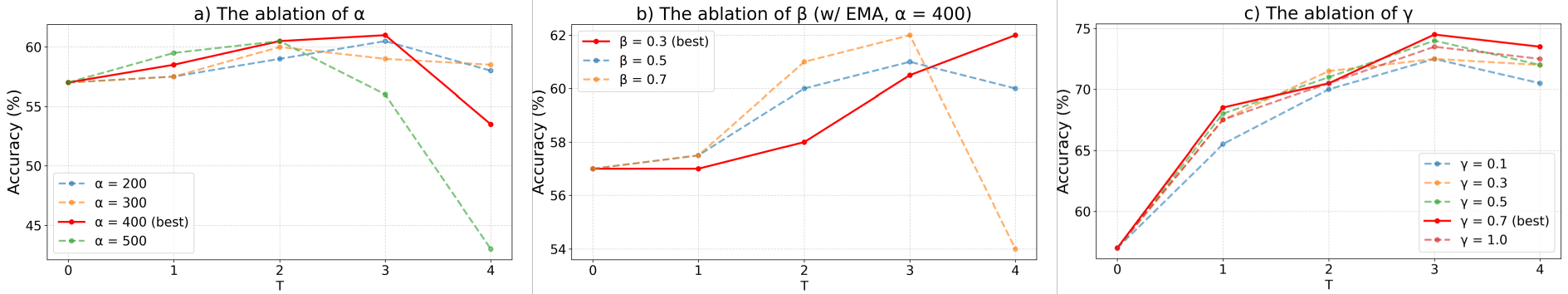}
  \caption{Ablation study of the ControlMLLM and ControlMLLM++. (a) Effect of different energy weights $\alpha$ of ControlMLLM on optimization stability and accuracy, larger $\alpha$ leading to overfitting of the model. (b) Influence of EMA decay rate $\beta$ of ControlMLLM (with fixed $\alpha = 400$), smaller value of $\beta$ results in slower convergence of the learnable latent variable. (c) Impact of language bias mitigation weight $\gamma$ in PromptDebias, $\gamma$=0.7 with the best performance.}

  \label{supfig:abl}
\end{figure*}

\begin{table*}[t]
\caption{Referring Description performance on RefCOCOg and Screenshot datasets. Metrics include BLEU-4 (B@4), METEOR (M), CIDEr (C), and SPICE (S). Our method not only equips non-referring models with effective grounding ability but also complements modern referring-capable MLLMs by enhancing their generalization and precision.}
\label{tab:pami_style_rd}
\centering
\begin{small}
\begin{sc}
\begin{tabular}{l|cccc|cccc}
\toprule
\multirow{2}{*}{Model} & \multicolumn{4}{c|}{RefCOCOg~(In-Domain)} & \multicolumn{4}{c}{Screenshot~(Out-of-Domain)} \\
 & B@4 & M & C & S & B@4 & M & C & S \\
\midrule
LLaVA-1.5 & \textcolor{gray}{5.02} & \textcolor{gray}{13.15} & \textcolor{gray}{55.61} & \textcolor{gray}{17.61} & \textcolor{gray}{0.32} & \textcolor{gray}{3.96} & \textcolor{gray}{9.80} & \textcolor{gray}{3.58} \\
LLaVA-1.5 + ControlMLLM & 5.53 & 14.00 & 59.75 & 19.08 & 0.45 & 5.08 & 19.74 & 5.81 \\
LLaVA-1.5 + ControlMLLM++ & \textbf{6.24} & \textbf{15.05} & \textbf{67.37} & \textbf{21.46} & \textbf{0.57} & \textbf{6.53} & \textbf{40.01} & \textbf{9.14} \\
\midrule
LLaVA-HR & \textcolor{gray}{5.28} & \textcolor{gray}{13.45} & \textcolor{gray}{56.29} & \textcolor{gray}{18.55} & \textcolor{gray}{0.29} & \textcolor{gray}{4.27} & \textcolor{gray}{10.88} & \textcolor{gray}{4.59} \\
LLaVA-HR + ControlMLLM & 6.32 & 15.00 & 68.82 & 21.55 & 0.64 & 6.79 & 37.10 & 8.54 \\
LLaVA-HR + ControlMLLM++ & \textbf{7.50} & \textbf{16.11} & \textbf{78.42} & \textbf{24.02} & \textbf{0.98} & \textbf{9.18} & \textbf{66.96} & \textbf{13.83} \\
\midrule
Qwen2.5-VL & 5.22 & 16.86 & 56.78 & 20.18 & 1.09 & 4.56 & 34.32 & 7.15 \\
Qwen2.5-VL + ControlMLLM & 5.33 & \textbf{16.91} & 58.20 & 20.12 & 4.26 & 9.91 & 86.35 & 15.27\\
Qwen2.5-VL + ControlMLLM++ & \textbf{5.45} & 16.53 & \textbf{59.50} & 19.95 & \textbf{9.05} & \textbf{16.04} & \textbf{141.36} & \textbf{25.08} \\
\bottomrule
\end{tabular}
\end{sc}
\end{small}
\end{table*}

\subsection{Ablation Study}
The ablation studies mainly focus on the box referred object classification task~(validation set for parameters ablation and test set for the ablation of modules).

\subsubsection{Ablation Study of ControlMLLM}
As shown in Figure~\ref{supfig:abl}~a), the larger $T$ results in a decrease in the model's accuracy on the ROC task~(validation set). And the value of $\alpha$ affects the convergence speed of optimization, with larger $\alpha$ also leading to overfitting of the model. And as shown in Figure~\ref{supfig:abl}~b), when equipped with a smaller $\beta$ value, it effectively mitigates the overfitting issue associated with the ControlMLLM. For instance, with $\alpha=400$ and $\beta=0.3$, the model's performance improves from 53.5 to 62.5. 
However, a smaller value of $\beta$ also results in slower convergence of the learnable latent variable. Therefore, we combine the Early Stop strategy, allowing us to use a slightly larger $\beta$ to accelerate the convergence of the learnable latent variable.
The experiment about ES is shown in Table~\ref{supTab2}. After incorporating the early stop strategy, 
we can opt for slightly larger $T$ to ensure that the model is adequately optimized on challenging samples. The early stop strategy allows us to attain superior model performance while reducing the impact of overfitting. 

\begin{table}[t]
\caption{The ablation of ES on ControlMLLM~($\alpha=400,\beta=0.5$) on validation set of ROC task.}
\label{supTab2}
\begin{small}
\begin{sc}
    \begin{tabular}{lcccc}
    \toprule
    $T\rightarrow$ & 0 & 4 & 5 \\
    \midrule
    Acc. & 57.00 & 60.50 & 62.50  \\
    \bottomrule
    \end{tabular}
\end{sc}
\end{small}
\end{table}

\begin{table}[t]
\caption{Ablation study of \textbf{ControlMLLM++} on the ROC task. \textbf{Optim++} includes three components: Adam optimizer, $A^{(st)}$ (answer-start token attention), and LayerSelection. \textbf{PromptDebias} denotes the language bias mitigation module. Accuracy (\textbf{Acc.}) is reported for each variant.}
\label{abl++}
\setlength{\tabcolsep}{4pt}
    \begin{tabular}{cccccc}
    \toprule
    \multicolumn{3}{c}{Optim++} & \multicolumn{1}{c}{} & \multicolumn{1}{c}{} \\
    \cmidrule(r){1-3}
    Adam & $A^{(st)}$ & LayerSelection  & PromptDebias & Acc. \\
    \midrule
     &  &  &  & 60.59 \\
    \checkmark &  &  &  & 66.93 \\
    \checkmark & \checkmark &  &  & 65.70 \\
    \checkmark &  & \checkmark &  & 65.70 \\
    \checkmark & \checkmark & \checkmark &  & 68.02 \\
    \checkmark & \checkmark & \checkmark & \checkmark & \textbf{71.19} \\
    \bottomrule
    \end{tabular}
\end{table}

\begin{table}[t]
\caption{
Performance comparison of Vanilla, ControlMLLM, and ControlMLLM++ across different MLLM architectures with different decoder sizes on the ROC task (box, test set).}

\label{supTab4}
\setlength{\tabcolsep}{3pt}
    \begin{tabular}{lccc}
    \toprule
     Models & Vanilla & ControlMLLM & ControlMLLM++  \\
    \midrule
     LLaVA-1.5-7B & \textcolor{gray}{54.72} & 60.59 & \textbf{71.19} \\
     LLaVA-1.5-13B & \textcolor{gray}{55.69} & 58.40 & \textbf{68.22} \\
     \midrule
    Qwen2.5-VL-3B &75.84 & 76.23  & \textbf{77.84} \\
    Qwen2.5-VL-7B & 78.81 & 79.20 & \textbf{79.20} \\
    \bottomrule
    \end{tabular}
\end{table}

\begin{table}[t]
\caption{Inference latency and GPU memory usage for ControlMLLM and ControlMLLM++ compared to the baseline LLaVA-1.5 model. Results are measured for short (6 tokens) and long (436 tokens) output sequences on a single NVIDIA GTX 4090 GPU. Early stopping is disabled in this experiment.}
\label{supTab3}
    \begin{tabular}{lcc}
    \toprule
    Models & Latency(s) & GPU Mem. \\
    \midrule
    LLaVA-1.5~(6 tokens) & 0.94 & 7.6G   \\
     +ControlMLLM & 1.88 & 13.6G \\
     +ControlMLLM++& 1.79 & 13.0G \\
     ~(w/o PromptDebias) & & \\
     +ControlMLLM++ & 2.54 & 21.3G \\
     \midrule
     LLaVA-1.5~(436 tokens) & 21.28 & 8.9G   \\
     +ControlMLLM & 22.22 & 13.6G \\
     +ControlMLLM++ & 21.99 & 13.0G \\
     ~(w/o PromptDebias) & & \\
     +ControlMLLM++ & 30.35 & 21.3G \\
    \bottomrule
    \end{tabular}
\end{table}

\subsubsection{Ablation Study of ControlMLLM++}
Table~\ref{abl++} presents an ablation study of the proposed ControlMLLM++ on the ROC benchmark, focusing on the impact of each component in the Optim++ as well as the PromptDebias. We decompose Optim++ into three key components: Adam optimizer, attention map restriction ($A^{(st)}$), and layer selection. As shown in the table, replacing the original SGD with Adam alone brings a notable performance gain (from 60.59 to 66.93), highlighting the benefit of adaptive optimization. Adding $A^{(st)}$, which restricts attention computation to the answer-start token, slightly decreases performance (66.93 → 65.70), suggesting that while it simplifies optimization, it may lose some information. Similarly, introducing layer selection alone achieves comparable performance (65.70), but combining both $A^{(st)}$ and layer selection further boosts performance to 68.02, indicating a synergistic effect in focusing optimization on more informative attention maps. Finally, integrating the PromptDebias strategy with Optim++ yields the best performance (71.84), demonstrating the effectiveness of jointly addressing both optimization stability and language bias.

\subsubsection{Impact of LLM Size on Different MLLMs}
Table~\ref{supTab4} examines the impact of decoder size on the ROC task performance across different MLLMs. We observe that increasing model size alone (e.g., from LLaVA-1.5-7B to 13B) yields only marginal improvements in the vanilla setting. In contrast, applying ControlMLLM and especially ControlMLLM++ brings substantial gains across all sizes. For example, ControlMLLM++ improves LLaVA-1.5-7B from 54.72 to 71.19, significantly outperforming the larger 13B baseline. Even for Qwen2.5-VL (from 3B to 7B), which already supports referring natively, ControlMLLM++ provides additional performance gains. These results indicate that our method complements model scaling and offers consistent benefits across model capacities.

\subsubsection{Inference Cost}
Table~\ref{supTab3} presents a comparison of inference latency and GPU memory usage among the baseline LLaVA-1.5 model, ControlMLLM, and our proposed ControlMLLM++. We evaluate both short (6 tokens) and long (436 tokens) output settings on a single NVIDIA RTX 4090 GPU with early stopping disabled. Compared to the baseline LLaVA-1.5, ControlMLLM introduces additional computation due to test-time optimization, leading to moderate increases in both latency (0.94s → 1.88s for short outputs) and memory consumption (7.6G → 13.6G). ControlMLLM++ (w/o PromptDebias) slightly improves inference efficiency over ControlMLLM, suggesting that our revised optimization strategy enhances stability and reduces redundant memory overhead. With the full version of ControlMLLM++, including PromptDebias, the latency and memory increase more significantly (2.54s and 21.3G for short outputs), primarily due to additional computation for extra decoding. This overhead is consistent in long output scenarios as well (e.g., 30.35s vs. 21.28s for LLaVA-1.5), reflecting a trade-off between performance gains and computational cost. Overall, while ControlMLLM++ incurs higher inference costs, it brings enhanced stability and improved alignment performance.


\section{Limitations}
\label{sec:limit}
While we have demonstrated visual prompt control by optimizing only visual tokens, our technique is subject to a few limitations. 
First, the gradient-based optimization introduces additional computational cost during inference. However, this can be significantly mitigated through engineering optimizations~(e.g., Ollama~\footnote{https://ollama.com/}) or efficient attention implementations. We consider this a reasonable trade-off for the gained control precision.
Second, our method currently requires access to model gradients and internal representations, limiting its application to open-source MLLMs. While this excludes proprietary black-box models, the growing ecosystem of powerful open-weight models ensures broad practical applicability.
Third, the current implementation supports control for only one image region at a time. Extending this to multi-region interaction presents interesting challenges in attention management and gradient conflict resolution, which we identify as a key direction for future research.

\section{Conclusion}
In this work, we present a test-time computing method, named as ControlMLLM, to integrate visual prompts into Multimodal Large Language Models (MLLMs) through learnable latent variable optimization. Based on ControlMLLM, we further introduce ControlMLLM++ which introduces a novel optimization strategy Optim++ that enhances training stability and a PromptDebias mechanism mitigates prompt language bias. By adjusting visual tokens during inference, our approach strengthens attention to referring regions, enabling detailed descriptions and reasoning without additional training costs. Our method supports various referring formats such as box, mask, scribble, and point. Experimental results demonstrate strong out-of-domain generalization and interpretability, making ControlMLLM++ a promising direction for embedding advanced referring abilities into MLLMs.

\section*{Acknowledgments}
This work was supported by the National Science Fund for Distinguished Young Scholars (No.62025603), the National Natural Science Foundation of China (No. U22B2051, No. 62302411) and China Postdoctoral Science Foundation (No. 2023M732948), and the Zhongguancun Academy, Beijing, China (No. 20240103).

\backmatter








\section*{Declarations}

\begin{itemize}
\item Funding: This work was supported by National Key R\&D Program of China (No.2023YFB4502804), the National Science Fund for Distinguished Young Scholars (No.62025603), the National Natural Science Foundation of China (No. U22B2051, No. U21B2037, No. 62302411), China Postdoctoral Science Foundation (No. 2023M732948), and the Zhongguancun Academy, Beijing, China (No. 20240103).
\item Conflict of interest/Competing interests: The authors have no relevant financial or non-financial interests to disclose.
\item Ethics approval and consent to participate: The authors have no relevant ethics approval to disclose.
\item Consent for publication: All authors agreed to publish the work.
\item Data availability: The dataset LVIS~\citep{gupta2019lvis} for this study can be downloaded at \url{https://www.lvisdataset.org}. The RefCOCOg~\citep{kazemzadeh2014referitgame} and Screenshot~\citep{cheng2024seeclick} can be downloaded at \url{https://huggingface.co/datasets/lmms-lab/RefCOCOg} and \url{https://huggingface.co/datasets/rootsautomation/ScreenSpot}.
\item Code availability: Code is made publicly available at: \url{https://github.com/mrwu-mac/ControlMLLM}.
\item Author contribution: Material preparation, data collection, and analysis were primarily conducted by Mingrui Wu, Hao Chen, and Jiayi Ji. The model was originally proposed by Mingrui Wu and Jiayi Ji, with improvements made by Hao Chen. Hao Chen was also responsible for experimental execution and plotting. Ming-Ming Cheng and Zhiyuan Liu contributed to experimental design and manuscript revisions. The project leaders, Rongrong Ji, Liujuan Cao, and Xiaoshuai Sun, played key roles in the feasibility discussions and manuscript refinement. Additionally, Xiaoshuai Sun contributed to revising the manuscript, providing critical insights into the model design, and contributing to the overall theoretical framework. The first draft of the manuscript was written by Mingrui Wu and Jiayi Ji, and all authors provided feedback on earlier versions. Finally, all authors read and approved the final manuscript.
\end{itemize}

\bibliographystyle{sn-basic}
\bibliography{neurips_2024}

@article{achiam2023gpt,
  title={Gpt-4 technical report},
  author={Achiam, Josh and Adler, Steven and Agarwal, Sandhini and Ahmad, Lama and Akkaya, Ilge and Aleman, Florencia Leoni and Almeida, Diogo and Altenschmidt, Janko and Altman, Sam and Anadkat, Shyamal and others},
  journal={arXiv preprint arXiv:2303.08774},
  year={2023}
}

@inproceedings{radford2021learning,
  title={Learning transferable visual models from natural language supervision},
  author={Radford, Alec and Kim, Jong Wook and Hallacy, Chris and Ramesh, Aditya and Goh, Gabriel and Agarwal, Sandhini and Sastry, Girish and Askell, Amanda and Mishkin, Pamela and Clark, Jack and others},
  booktitle={International conference on machine learning},
  pages={8748--8763},
  year={2021},
  organization={PMLR}
}

@article{liu2024visual,
  title={Visual instruction tuning},
  author={Liu, Haotian and Li, Chunyuan and Wu, Qingyang and Lee, Yong Jae},
  journal={Advances in neural information processing systems},
  volume={36},
  year={2024}
}

@article{ye2023mplug,
  title={mplug-owl2: Revolutionizing multi-modal large language model with modality collaboration},
  author={Ye, Qinghao and Xu, Haiyang and Ye, Jiabo and Yan, Ming and Liu, Haowei and Qian, Qi and Zhang, Ji and Huang, Fei and Zhou, Jingren},
  journal={arXiv preprint arXiv:2311.04257},
  year={2023}
}

@article{zhu2023minigpt,
  title={Minigpt-4: Enhancing vision-language understanding with advanced large language models},
  author={Zhu, Deyao and Chen, Jun and Shen, Xiaoqian and Li, Xiang and Elhoseiny, Mohamed},
  journal={arXiv preprint arXiv:2304.10592},
  year={2023}
}

@inproceedings{li2023blip,
  title={Blip-2: Bootstrapping language-image pre-training with frozen image encoders and large language models},
  author={Li, Junnan and Li, Dongxu and Savarese, Silvio and Hoi, Steven},
  booktitle={International conference on machine learning},
  pages={19730--19742},
  year={2023},
  organization={PMLR}
}

@article{dai2024instructblip,
  title={Instructblip: Towards general-purpose vision-language models with instruction tuning},
  author={Dai, Wenliang and Li, Junnan and Li, Dongxu and Tiong, Anthony Meng Huat and Zhao, Junqi and Wang, Weisheng and Li, Boyang and Fung, Pascale N and Hoi, Steven},
  journal={Advances in Neural Information Processing Systems},
  volume={36},
  year={2024}
}

@article{luo2024cheap,
  title={Cheap and quick: Efficient vision-language instruction tuning for large language models},
  author={Luo, Gen and Zhou, Yiyi and Ren, Tianhe and Chen, Shengxin and Sun, Xiaoshuai and Ji, Rongrong},
  journal={Advances in Neural Information Processing Systems},
  volume={36},
  year={2024}
}

@article{luo2024feast,
  title={Feast Your Eyes: Mixture-of-Resolution Adaptation for Multimodal Large Language Models},
  author={Luo, Gen and Zhou, Yiyi and Zhang, Yuxin and Zheng, Xiawu and Sun, Xiaoshuai and Ji, Rongrong},
  journal={arXiv preprint arXiv:2403.03003},
  year={2024}
}

@article{li2023otter,
  title={Otter: A Multi-Modal Model with In-Context Instruction Tuning},
  author={Li, Bo and Zhang, Yuanhan and Chen, Liangyu and Wang, Jinghao and Yang, Jingkang and Liu, Ziwei},
  journal={arXiv preprint arXiv:2305.03726},
  year={2023}
}

@article{internlmxcomposer2_4khd,
      title={InternLM-XComposer2-4KHD: A Pioneering Large Vision-Language Model Handling Resolutions from 336 Pixels to 4K HD},
      author={Xiaoyi Dong and Pan Zhang and Yuhang Zang and Yuhang Cao and Bin Wang and Linke Ouyang and Songyang Zhang and Haodong Duan and Wenwei Zhang and Yining Li and Hang Yan and Yang Gao and Zhe Chen and Xinyue Zhang and Wei Li and Jingwen Li and Wenhai Wang and Kai Chen and Conghui He and Xingcheng Zhang and Jifeng Dai and Yu Qiao and Dahua Lin and Jiaqi Wang},
      journal={arXiv preprint arXiv:2404.06512},
      year={2024}
}

@article{gao2023llamaadapterv2,
  title = {LLaMA-Adapter V2: Parameter-Efficient Visual Instruction Model},
  author={Gao, Peng and Han, Jiaming and Zhang, Renrui and Lin, Ziyi and Geng, Shijie and Zhou, Aojun and Zhang, Wei and Lu, Pan and He, Conghui and Yue, Xiangyu and Li, Hongsheng and Qiao, Yu},
  journal={arXiv preprint arXiv:2304.15010},
  year={2023}
}

@article{chen2023internvl,
  title={Internvl: Scaling up vision foundation models and aligning for generic visual-linguistic tasks},
  author={Chen, Zhe and Wu, Jiannan and Wang, Wenhai and Su, Weijie and Chen, Guo and Xing, Sen and Muyan, Zhong and Zhang, Qinglong and Zhu, Xizhou and Lu, Lewei and others},
  journal={arXiv preprint arXiv:2312.14238},
  year={2023}
}

@article{liu2023improved,
  title={Improved baselines with visual instruction tuning},
  author={Liu, Haotian and Li, Chunyuan and Li, Yuheng and Lee, Yong Jae},
  journal={arXiv preprint arXiv:2310.03744},
  year={2023}
}

@article{wang2023cogvlm,
  title={Cogvlm: Visual expert for pretrained language models},
  author={Wang, Weihan and Lv, Qingsong and Yu, Wenmeng and Hong, Wenyi and Qi, Ji and Wang, Yan and Ji, Junhui and Yang, Zhuoyi and Zhao, Lei and Song, Xixuan and others},
  journal={arXiv preprint arXiv:2311.03079},
  year={2023}
}

@article{touvron2023llama,
  title={Llama: Open and efficient foundation language models},
  author={Touvron, Hugo and Lavril, Thibaut and Izacard, Gautier and Martinet, Xavier and Lachaux, Marie-Anne and Lacroix, Timoth{\'e}e and Rozi{\`e}re, Baptiste and Goyal, Naman and Hambro, Eric and Azhar, Faisal and others},
  journal={arXiv preprint arXiv:2302.13971},
  year={2023}
}

@misc{vicuna2023,
    title = {Vicuna: An Open-Source Chatbot Impressing GPT-4 with 90\%* ChatGPT Quality},
    url = {https://lmsys.org/blog/2023-03-30-vicuna/},
    author = {Chiang, Wei-Lin and Li, Zhuohan and Lin, Zi and Sheng, Ying and Wu, Zhanghao and Zhang, Hao and Zheng, Lianmin and Zhuang, Siyuan and Zhuang, Yonghao and Gonzalez, Joseph E. and Stoica, Ion and Xing, Eric P.},
    month = {March},
    year = {2023}
}

@article{vaswani2017attention,
  title={Attention is all you need},
  author={Vaswani, Ashish and Shazeer, Noam and Parmar, Niki and Uszkoreit, Jakob and Jones, Llion and Gomez, Aidan N and Kaiser, {\L}ukasz and Polosukhin, Illia},
  journal={Advances in neural information processing systems},
  volume={30},
  year={2017}
}

@article{zhang2024ferret,
  title={Ferret-v2: An Improved Baseline for Referring and Grounding with Large Language Models},
  author={Zhang, Haotian and You, Haoxuan and Dufter, Philipp and Zhang, Bowen and Chen, Chen and Chen, Hong-You and Fu, Tsu-Jui and Wang, William Yang and Chang, Shih-Fu and Gan, Zhe and others},
  journal={arXiv preprint arXiv:2404.07973},
  year={2024}
}

@article{you2023ferret,
  title={Ferret: Refer and ground anything anywhere at any granularity},
  author={You, Haoxuan and Zhang, Haotian and Gan, Zhe and Du, Xianzhi and Zhang, Bowen and Wang, Zirui and Cao, Liangliang and Chang, Shih-Fu and Yang, Yinfei},
  journal={arXiv preprint arXiv:2310.07704},
  year={2023}
}

@article{zhang2023gpt4roi,
  title={Gpt4roi: Instruction tuning large language model on region-of-interest},
  author={Zhang, Shilong and Sun, Peize and Chen, Shoufa and Xiao, Min and Shao, Wenqi and Zhang, Wenwei and Chen, Kai and Luo, Ping},
  journal={arXiv preprint arXiv:2307.03601},
  year={2023}
}

@article{bai2023qwen,
  title={Qwen-vl: A frontier large vision-language model with versatile abilities},
  author={Bai, Jinze and Bai, Shuai and Yang, Shusheng and Wang, Shijie and Tan, Sinan and Wang, Peng and Lin, Junyang and Zhou, Chang and Zhou, Jingren},
  journal={arXiv preprint arXiv:2308.12966},
  year={2023}
}

@article{chen2023shikra,
  title={Shikra: Unleashing Multimodal LLM's Referential Dialogue Magic},
  author={Chen, Keqin and Zhang, Zhao and Zeng, Weili and Zhang, Richong and Zhu, Feng and Zhao, Rui},
  journal={arXiv preprint arXiv:2306.15195},
  year={2023}
}

@article{yue2024sc,
  title={SC-Tune: Unleashing Self-Consistent Referential Comprehension in Large Vision Language Models},
  author={Yue, Tongtian and Cheng, Jie and Guo, Longteng and Dai, Xingyuan and Zhao, Zijia and He, Xingjian and Xiong, Gang and Lv, Yisheng and Liu, Jing},
  journal={arXiv preprint arXiv:2403.13263},
  year={2024}
}

@article{lu2023lyrics,
  title={Lyrics: Boosting Fine-grained Language-Vision Alignment and Comprehension via Semantic-aware Visual Objects},
  author={Lu, Junyu and Gan, Ruyi and Zhang, Dixiang and Wu, Xiaojun and Wu, Ziwei and Sun, Renliang and Zhang, Jiaxing and Zhang, Pingjian and Song, Yan},
  journal={arXiv preprint arXiv:2312.05278},
  year={2023}
}

@article{peng2023kosmos,
  title={Kosmos-2: Grounding multimodal large language models to the world},
  author={Peng, Zhiliang and Wang, Wenhui and Dong, Li and Hao, Yaru and Huang, Shaohan and Ma, Shuming and Wei, Furu},
  journal={arXiv preprint arXiv:2306.14824},
  year={2023}
}

@article{zhang2023llava,
  title={Llava-grounding: Grounded visual chat with large multimodal models},
  author={Zhang, Hao and Li, Hongyang and Li, Feng and Ren, Tianhe and Zou, Xueyan and Liu, Shilong and Huang, Shijia and Gao, Jianfeng and Zhang, Lei and Li, Chunyuan and others},
  journal={arXiv preprint arXiv:2312.02949},
  year={2023}
}

@article{ma2024groma,
  title={Groma: Localized Visual Tokenization for Grounding Multimodal Large Language Models},
  author={Ma, Chuofan and Jiang, Yi and Wu, Jiannan and Yuan, Zehuan and Qi, Xiaojuan},
  journal={arXiv preprint arXiv:2404.13013},
  year={2024}
}

@article{you2024ferret,
  title={Ferret-UI: Grounded Mobile UI Understanding with Multimodal LLMs},
  author={You, Keen and Zhang, Haotian and Schoop, Eldon and Weers, Floris and Swearngin, Amanda and Nichols, Jeffrey and Yang, Yinfei and Gan, Zhe},
  journal={arXiv preprint arXiv:2404.05719},
  year={2024}
}

@article{he2024multi,
  title={Multi-modal Instruction Tuned LLMs with Fine-grained Visual Perception},
  author={He, Junwen and Wang, Yifan and Wang, Lijun and Lu, Huchuan and He, Jun-Yan and Lan, Jin-Peng and Luo, Bin and Xie, Xuansong},
  journal={arXiv preprint arXiv:2403.02969},
  year={2024}
}

@article{xuan2023pink,
  title={Pink: Unveiling the power of referential comprehension for multi-modal llms},
  author={Xuan, Shiyu and Guo, Qingpei and Yang, Ming and Zhang, Shiliang},
  journal={arXiv preprint arXiv:2310.00582},
  year={2023}
}

@article{lin2024draw,
  title={Draw-and-Understand: Leveraging Visual Prompts to Enable MLLMs to Comprehend What You Want},
  author={Lin, Weifeng and Wei, Xinyu and An, Ruichuan and Gao, Peng and Zou, Bocheng and Luo, Yulin and Huang, Siyuan and Zhang, Shanghang and Li, Hongsheng},
  journal={arXiv preprint arXiv:2403.20271},
  year={2024}
}

@article{yuan2023osprey,
  title={Osprey: Pixel Understanding with Visual Instruction Tuning},
  author={Yuan, Yuqian and Li, Wentong and Liu, Jian and Tang, Dongqi and Luo, Xinjie and Qin, Chi and Zhang, Lei and Zhu, Jianke},
  journal={arXiv preprint arXiv:2312.10032},
  year={2023}
}

@article{zhao2023chatspot,
  title={Chatspot: Bootstrapping multimodal llms via precise referring instruction tuning},
  author={Zhao, Liang and Yu, En and Ge, Zheng and Yang, Jinrong and Wei, Haoran and Zhou, Hongyu and Sun, Jianjian and Peng, Yuang and Dong, Runpei and Han, Chunrui and others},
  journal={arXiv preprint arXiv:2307.09474},
  year={2023}
}

@article{chen2023position,
  title={Position-enhanced visual instruction tuning for multimodal large language models},
  author={Chen, Chi and Qin, Ruoyu and Luo, Fuwen and Mi, Xiaoyue and Li, Peng and Sun, Maosong and Liu, Yang},
  journal={arXiv preprint arXiv:2308.13437},
  year={2023}
}

@article{rasheed2023glamm,
  title={Glamm: Pixel grounding large multimodal model},
  author={Rasheed, Hanoona and Maaz, Muhammad and Shaji, Sahal and Shaker, Abdelrahman and Khan, Salman and Cholakkal, Hisham and Anwer, Rao M and Xing, Erix and Yang, Ming-Hsuan and Khan, Fahad S},
  journal={arXiv preprint arXiv:2311.03356},
  year={2023}
}

@article{zhou2023regionblip,
  title={Regionblip: A unified multi-modal pre-training framework for holistic and regional comprehension},
  author={Zhou, Qiang and Yu, Chaohui and Zhang, Shaofeng and Wu, Sitong and Wang, Zhibing and Wang, Fan},
  journal={arXiv preprint arXiv:2308.02299},
  year={2023}
}

@article{xu2023pixel,
  title={Pixel aligned language models},
  author={Xu, Jiarui and Zhou, Xingyi and Yan, Shen and Gu, Xiuye and Arnab, Anurag and Sun, Chen and Wang, Xiaolong and Schmid, Cordelia},
  journal={arXiv preprint arXiv:2312.09237},
  year={2023}
}

@article{cai2023making,
  title={Making large multimodal models understand arbitrary visual prompts},
  author={Cai, Mu and Liu, Haotian and Mustikovela, Siva Karthik and Meyer, Gregory P and Chai, Yuning and Park, Dennis and Lee, Yong Jae},
  journal={arXiv preprint arXiv:2312.00784},
  year={2023}
}

@article{guo2024regiongpt,
  title={Regiongpt: Towards region understanding vision language model},
  author={Guo, Qiushan and De Mello, Shalini and Yin, Hongxu and Byeon, Wonmin and Cheung, Ka Chun and Yu, Yizhou and Luo, Ping and Liu, Sifei},
  journal={arXiv preprint arXiv:2403.02330},
  year={2024}
}

@article{ranasinghe2024learning,
  title={Learning to localize objects improves spatial reasoning in visual-llms},
  author={Ranasinghe, Kanchana and Shukla, Satya Narayan and Poursaeed, Omid and Ryoo, Michael S and Lin, Tsung-Yu},
  journal={arXiv preprint arXiv:2404.07449},
  year={2024}
}

@article{tian2024chatterbox,
  title={ChatterBox: Multi-round Multimodal Referring and Grounding},
  author={Tian, Yunjie and Ma, Tianren and Xie, Lingxi and Qiu, Jihao and Tang, Xi and Zhang, Yuan and Jiao, Jianbin and Tian, Qi and Ye, Qixiang},
  journal={arXiv preprint arXiv:2401.13307},
  year={2024}
}

@article{zhan2024griffon,
  title={Griffon v2: Advancing multimodal perception with high-resolution scaling and visual-language co-referring},
  author={Zhan, Yufei and Zhu, Yousong and Zhao, Hongyin and Yang, Fan and Tang, Ming and Wang, Jinqiao},
  journal={arXiv preprint arXiv:2403.09333},
  year={2024}
}

@article{heo2025omni,
  title={Omni-RGPT: Unifying Image and Video Region-level Understanding via Token Marks},
  author={Heo, Miran and Chen, Min-Hung and Huang, De-An and Liu, Sifei and Radhakrishnan, Subhashree and Kim, Seon Joo and Wang, Yu-Chiang Frank and Hachiuma, Ryo},
  journal={arXiv preprint arXiv:2501.08326},
  year={2025}
}

@inproceedings{shtedritski2023does,
  title={What does clip know about a red circle? visual prompt engineering for vlms},
  author={Shtedritski, Aleksandar and Rupprecht, Christian and Vedaldi, Andrea},
  booktitle={Proceedings of the IEEE/CVF International Conference on Computer Vision},
  pages={11987--11997},
  year={2023}
}

@article{yang2024fine,
  title={Fine-grained visual prompting},
  author={Yang, Lingfeng and Wang, Yueze and Li, Xiang and Wang, Xinlong and Yang, Jian},
  journal={Advances in Neural Information Processing Systems},
  volume={36},
  year={2024}
}

@article{yang2023set,
  title={Set-of-mark prompting unleashes extraordinary visual grounding in gpt-4v},
  author={Yang, Jianwei and Zhang, Hao and Li, Feng and Zou, Xueyan and Li, Chunyuan and Gao, Jianfeng},
  journal={arXiv preprint arXiv:2310.11441},
  year={2023}
}

@article{sun2023alpha,
  title={Alpha-CLIP: A clip model focusing on wherever you want},
  author={Sun, Zeyi and Fang, Ye and Wu, Tong and Zhang, Pan and Zang, Yuhang and Kong, Shu and Xiong, Yuanjun and Lin, Dahua and Wang, Jiaqi},
  journal={arXiv preprint arXiv:2312.03818},
  year={2023}
}

@article{zhang2023prompt,
  title={Prompt Highlighter: Interactive Control for Multi-Modal LLMs},
  author={Zhang, Yuechen and Qian, Shengju and Peng, Bohao and Liu, Shu and Jia, Jiaya},
  journal={arXiv preprint arXiv:2312.04302},
  year={2023}
}

@article{wang2023caption,
  title={Caption anything: Interactive image description with diverse multimodal controls},
  author={Wang, Teng and Zhang, Jinrui and Fei, Junjie and Ge, Yixiao and Zheng, Hao and Tang, Yunlong and Li, Zhe and Gao, Mingqi and Zhao, Shanshan and Shan, Ying and others},
  journal={arXiv preprint arXiv:2305.02677},
  year={2023}
}

@inproceedings{jia2022visual,
  title={Visual prompt tuning},
  author={Jia, Menglin and Tang, Luming and Chen, Bor-Chun and Cardie, Claire and Belongie, Serge and Hariharan, Bharath and Lim, Ser-Nam},
  booktitle={European Conference on Computer Vision},
  pages={709--727},
  year={2022},
  organization={Springer}
}

@article{bahng2022exploring,
  title={Exploring visual prompts for adapting large-scale models},
  author={Bahng, Hyojin and Jahanian, Ali and Sankaranarayanan, Swami and Isola, Phillip},
  journal={arXiv preprint arXiv:2203.17274},
  year={2022}
}

@article{zhang2024exploring,
  title={Exploring the Transferability of Visual Prompting for Multimodal Large Language Models},
  author={Zhang, Yichi and Dong, Yinpeng and Zhang, Siyuan and Min, Tianzan and Su, Hang and Zhu, Jun},
  journal={arXiv preprint arXiv:2404.11207},
  year={2024}
}

@article{shu2022test,
  title={Test-time prompt tuning for zero-shot generalization in vision-language models},
  author={Shu, Manli and Nie, Weili and Huang, De-An and Yu, Zhiding and Goldstein, Tom and Anandkumar, Anima and Xiao, Chaowei},
  journal={Advances in Neural Information Processing Systems},
  volume={35},
  pages={14274--14289},
  year={2022}
}

@inproceedings{wu2022difnet,
  title={Difnet: Boosting visual information flow for image captioning},
  author={Wu, Mingrui and Zhang, Xuying and Sun, Xiaoshuai and Zhou, Yiyi and Chen, Chao and Gu, Jiaxin and Sun, Xing and Ji, Rongrong},
  booktitle={Proceedings of the IEEE/CVF conference on computer vision and pattern recognition},
  pages={18020--18029},
  year={2022}
}

@inproceedings{chen2024training,
  title={Training-free layout control with cross-attention guidance},
  author={Chen, Minghao and Laina, Iro and Vedaldi, Andrea},
  booktitle={Proceedings of the IEEE/CVF Winter Conference on Applications of Computer Vision},
  pages={5343--5353},
  year={2024}
}

@article{hertz2022prompt,
  title={Prompt-to-prompt image editing with cross attention control},
  author={Hertz, Amir and Mokady, Ron and Tenenbaum, Jay and Aberman, Kfir and Pritch, Yael and Cohen-Or, Daniel},
  journal={arXiv preprint arXiv:2208.01626},
  year={2022}
}

@inproceedings{xie2023boxdiff,
  title={Boxdiff: Text-to-image synthesis with training-free box-constrained diffusion},
  author={Xie, Jinheng and Li, Yuexiang and Huang, Yawen and Liu, Haozhe and Zhang, Wentian and Zheng, Yefeng and Shou, Mike Zheng},
  booktitle={Proceedings of the IEEE/CVF International Conference on Computer Vision},
  pages={7452--7461},
  year={2023}
}

@inproceedings{kim2023dense,
  title={Dense text-to-image generation with attention modulation},
  author={Kim, Yunji and Lee, Jiyoung and Kim, Jin-Hwa and Ha, Jung-Woo and Zhu, Jun-Yan},
  booktitle={Proceedings of the IEEE/CVF International Conference on Computer Vision},
  pages={7701--7711},
  year={2023}
}

@misc{rombach2021highresolution,
      title={High-Resolution Image Synthesis with Latent Diffusion Models}, 
      author={Robin Rombach and Andreas Blattmann and Dominik Lorenz and Patrick Esser and Björn Ommer},
      year={2021},
      eprint={2112.10752},
      archivePrefix={arXiv},
      primaryClass={cs.CV}
}

@article{zhu2024ibd,
  title={IBD: Alleviating Hallucinations in Large Vision-Language Models via Image-Biased Decoding},
  author={Zhu, Lanyun and Ji, Deyi and Chen, Tianrun and Xu, Peng and Ye, Jieping and Liu, Jun},
  journal={arXiv preprint arXiv:2402.18476},
  year={2024}
}

@article{chen2024image,
  title={An Image is Worth 1/2 Tokens After Layer 2: Plug-and-Play Inference Acceleration for Large Vision-Language Models},
  author={Chen, Liang and Zhao, Haozhe and Liu, Tianyu and Bai, Shuai and Lin, Junyang and Zhou, Chang and Chang, Baobao},
  journal={arXiv preprint arXiv:2403.06764},
  year={2024}
}

@inproceedings{kirillov2023segment,
  title={Segment anything},
  author={Kirillov, Alexander and Mintun, Eric and Ravi, Nikhila and Mao, Hanzi and Rolland, Chloe and Gustafson, Laura and Xiao, Tete and Whitehead, Spencer and Berg, Alexander C and Lo, Wan-Yen and others},
  booktitle={Proceedings of the IEEE/CVF International Conference on Computer Vision},
  pages={4015--4026},
  year={2023}
}

@inproceedings{gupta2019lvis,
  title={{LVIS}: A Dataset for Large Vocabulary Instance Segmentation},
  author={Gupta, Agrim and Dollar, Piotr and Girshick, Ross},
  booktitle={Proceedings of the {IEEE} Conference on Computer Vision and Pattern Recognition},
  year={2019}
}

@article{bradski2000opencv,
  title={The opencv library.},
  author={Bradski, Gary},
  journal={Dr. Dobb's Journal: Software Tools for the Professional Programmer},
  volume={25},
  number={11},
  pages={120--123},
  year={2000},
  publisher={Miller Freeman Inc.}
}

@inproceedings{kazemzadeh2014referitgame,
  title={Referitgame: Referring to objects in photographs of natural scenes},
  author={Kazemzadeh, Sahar and Ordonez, Vicente and Matten, Mark and Berg, Tamara},
  booktitle={Proceedings of the 2014 conference on empirical methods in natural language processing (EMNLP)},
  pages={787--798},
  year={2014}
}

@inproceedings{papineni2002bleu,
  title={Bleu: a method for automatic evaluation of machine translation},
  author={Papineni, Kishore and Roukos, Salim and Ward, Todd and Zhu, Wei-Jing},
  booktitle={Proceedings of the 40th annual meeting of the Association for Computational Linguistics},
  pages={311--318},
  year={2002}
}

@inproceedings{vedantam2015cider,
  title={Cider: Consensus-based image description evaluation},
  author={Vedantam, Ramakrishna and Lawrence Zitnick, C and Parikh, Devi},
  booktitle={Proceedings of the IEEE conference on computer vision and pattern recognition},
  pages={4566--4575},
  year={2015}
}

@inproceedings{banerjee2005meteor,
  title={METEOR: An automatic metric for MT evaluation with improved correlation with human judgments},
  author={Banerjee, Satanjeev and Lavie, Alon},
  booktitle={Proceedings of the acl workshop on intrinsic and extrinsic evaluation measures for machine translation and/or summarization},
  pages={65--72},
  year={2005}
}

@inproceedings{anderson2016spice,
  title={Spice: Semantic propositional image caption evaluation},
  author={Anderson, Peter and Fernando, Basura and Johnson, Mark and Gould, Stephen},
  booktitle={Computer Vision--ECCV 2016: 14th European Conference, Amsterdam, The Netherlands, October 11-14, 2016, Proceedings, Part V 14},
  pages={382--398},
  year={2016},
  organization={Springer}
}

@article{veit2016coco,
  title={Coco-text: Dataset and benchmark for text detection and recognition in natural images},
  author={Veit, Andreas and Matera, Tomas and Neumann, Lukas and Matas, Jiri and Belongie, Serge},
  journal={arXiv preprint arXiv:1601.07140},
  year={2016}
}

@article{oquab2023dinov2,
  title={Dinov2: Learning robust visual features without supervision},
  author={Oquab, Maxime and Darcet, Timoth{\'e}e and Moutakanni, Th{\'e}o and Vo, Huy and Szafraniec, Marc and Khalidov, Vasil and Fernandez, Pierre and Haziza, Daniel and Massa, Francisco and El-Nouby, Alaaeldin and others},
  journal={arXiv preprint arXiv:2304.07193},
  year={2023}
}

@inproceedings{zhang2021rstnet,
  title={Rstnet: Captioning with adaptive attention on visual and non-visual words},
  author={Zhang, Xuying and Sun, Xiaoshuai and Luo, Yunpeng and Ji, Jiayi and Zhou, Yiyi and Wu, Yongjian and Huang, Feiyue and Ji, Rongrong},
  booktitle={Proceedings of the IEEE/CVF conference on computer vision and pattern recognition},
  pages={15465--15474},
  year={2021}
}

@misc{fei2024vitron,
  title={VITRON: A Unified Pixel-level Vision LLM for Understanding, Generating, Segmenting, Editing},
  author={Fei, Hao and Wu, Shengqiong and Zhang, Hanwang and Chua, Tat-Seng and Yan, Shuicheng},
  year={2024},
  publisher={CoRR}
}

@article{fei2024enhancing,
  title={Enhancing video-language representations with structural spatio-temporal alignment},
  author={Fei, Hao and Wu, Shengqiong and Zhang, Meishan and Zhang, Min and Chua, Tat-Seng and Yan, Shuicheng},
  journal={IEEE Transactions on Pattern Analysis and Machine Intelligence},
  year={2024},
  publisher={IEEE}
}

@inproceedings{fei2024video,
  title={Video-of-thought: Step-by-step video reasoning from perception to cognition},
  author={Fei, Hao and Wu, Shengqiong and Ji, Wei and Zhang, Hanwang and Zhang, Meishan and Lee, Mong-Li and Hsu, Wynne},
  booktitle={Forty-first International Conference on Machine Learning},
  year={2024}
}

@inproceedings{gong2022person,
  title={Person re-identification method based on color attack and joint defence},
  author={Gong, Yunpeng and Huang, Liqing and Chen, Lifei},
  booktitle={Proceedings of the IEEE/CVF conference on computer vision and pattern recognition},
  pages={4313--4322},
  year={2022}
}

@InProceedings{pmlr-v235-wu24l,
  title = 	 {Evaluating and Analyzing Relationship Hallucinations in Large Vision-Language Models},
  author =       {Wu, Mingrui and Ji, Jiayi and Huang, Oucheng and Li, Jiale and Wu, Yuhang and Sun, Xiaoshuai and Ji, Rongrong},
  booktitle = 	 {Proceedings of the 41st International Conference on Machine Learning},
  pages = 	 {53553--53570},
  year = 	 {2024},
  volume = 	 {235},
  series = 	 {Proceedings of Machine Learning Research},
  month = 	 {21--27 Jul},
  publisher =    {PMLR},
  url = 	 {https://proceedings.mlr.press/v235/wu24l.html},
  }

@article{wu2024tradiffusion,
  title={TraDiffusion: Trajectory-Based Training-Free Image Generation},
  author={Wu, Mingrui and Huang, Oucheng and Ji, Jiayi and Li, Jiale and Cai, Xinyue and Kuang, Huafeng and Liu, Jianzhuang and Sun, Xiaoshuai and Ji, Rongrong},
  journal={arXiv preprint arXiv:2408.09739},
  year={2024}
}

@article{yao2024cpt,
  title={Cpt: Colorful prompt tuning for pre-trained vision-language models},
  author={Yao, Yuan and Zhang, Ao and Zhang, Zhengyan and Liu, Zhiyuan and Chua, Tat-Seng and Sun, Maosong},
  journal={AI Open},
  volume={5},
  pages={30--38},
  year={2024},
  publisher={Elsevier}
}

@article{hu2024minicpm,
  title={Minicpm: Unveiling the potential of small language models with scalable training strategies},
  author={Hu, Shengding and Tu, Yuge and Han, Xu and He, Chaoqun and Cui, Ganqu and Long, Xiang and Zheng, Zhi and Fang, Yewei and Huang, Yuxiang and Zhao, Weilin and others},
  journal={arXiv preprint arXiv:2404.06395},
  year={2024}
}

@inproceedings{leng2024mitigating,
  title={Mitigating object hallucinations in large vision-language models through visual contrastive decoding},
  author={Leng, Sicong and Zhang, Hang and Chen, Guanzheng and Li, Xin and Lu, Shijian and Miao, Chunyan and Bing, Lidong},
  booktitle={Proceedings of the IEEE/CVF Conference on Computer Vision and Pattern Recognition},
  pages={13872--13882},
  year={2024}
}

@article{zhang2025mllms,
  title={MLLMs know where to look: Training-free perception of small visual details with multimodal LLMs},
  author={Zhang, Jiarui and Khayatkhoei, Mahyar and Chhikara, Prateek and Ilievski, Filip},
  journal={arXiv preprint arXiv:2502.17422},
  year={2025}
}

@article{lim2025ureca,
  title={URECA: Unique Region Caption Anything},
  author={Lim, Sangbeom and Kim, Junwan and Yoon, Heeji and Jung, Jaewoo and Kim, Seungryong},
  journal={arXiv preprint arXiv:2504.05305},
  year={2025}
}

@article{wu2024accelerating,
  title={Accelerating multimodal large language models via dynamic visual-token exit and the empirical findings},
  author={Wu, Qiong and Lin, Wenhao and Ye, Weihao and Zhou, Yiyi and Sun, Xiaoshuai and Ji, Rongrong},
  journal={arXiv preprint arXiv:2411.19628},
  year={2024}
}

@article{bai2025qwen2,
  title={Qwen2. 5-vl technical report},
  author={Bai, Shuai and Chen, Keqin and Liu, Xuejing and Wang, Jialin and Ge, Wenbin and Song, Sibo and Dang, Kai and Wang, Peng and Wang, Shijie and Tang, Jun and others},
  journal={arXiv preprint arXiv:2502.13923},
  year={2025}
}

@article{zhu2025internvl3,
  title={InternVL3: Exploring Advanced Training and Test-Time Recipes for Open-Source Multimodal Models},
  author={Zhu, Jinguo and Wang, Weiyun and Chen, Zhe and Liu, Zhaoyang and Ye, Shenglong and Gu, Lixin and Duan, Yuchen and Tian, Hao and Su, Weijie and Shao, Jie and others},
  journal={arXiv preprint arXiv:2504.10479},
  year={2025}
}

@misc{cheng2024seeclick,
      title={SeeClick: Harnessing GUI Grounding for Advanced Visual GUI Agents}, 
      author={Kanzhi Cheng and Qiushi Sun and Yougang Chu and Fangzhi Xu and Yantao Li and Jianbing Zhang and Zhiyong Wu},
      year={2024},
      eprint={2401.10935},
      archivePrefix={arXiv},
      primaryClass={cs.HC}
}

\end{document}